%% file: main.tex
\documentclass[preprint,12pt]{elsarticle}

\usepackage[T1]{fontenc}
\usepackage[utf8]{inputenc}
\usepackage{amsmath,amssymb}
\usepackage{graphicx}
\usepackage{booktabs}
\usepackage{siunitx}
\usepackage{tabularx}
\DeclareSIUnit{\bpm}{bpm}
\DeclareSIUnit{\mmHg}{mmHg}
\usepackage{xcolor}
\usepackage{url}
\graphicspath{{figures/}}

\newcommand{\SuppFigConsort}{Supplementary Figure~S1} \newcommand{\SuppFigPipeline}{Supplementary Figure~S2} \newcommand{\SuppTabBaseline}{Supplementary Table~S1} \newcommand{\SuppTabStateVariables}{Supplementary Table~S2} \newcommand{\SuppTabActions}{Supplementary Table~S3} \newcommand{\SuppTabCandidateSets}{Supplementary Table~S4}

\begin{document}

%% ---------------------------------------------------------------------
%% FRONT MATTER
%% ---------------------------------------------------------------------
\begin{frontmatter}

%% TITLE — provisional; refine once narrative is fixed.
\title{Offline Reinforcement Learning for Hemodynamic Management of
Sepsis in the ICU: a MIMIC-IV Study with Dual Off-Policy Evaluation}

%% AUTHORS — Marc Pérez Roig (author), David Fernández-Narro and Carlos Sáez
%% Silvestre (project directors / supervisors). AIME forbids authorship changes after
%% acceptance, so confirm order and affiliation details before submission.
\author[aff1]{Marc P\'erez-Roig}
\ead{perezroig.marc.dev@gmail.com}
\author[aff1]{David Fern\'andez-Narro}
\author[aff1]{Carlos S\'aez\corref{cor1}}
\ead{carsaesi@upv.es}
\cortext[cor1]{Corresponding author.}
\affiliation[aff1]{organization={Biomedical Data Science Lab, Instituto Universitario de
  Tecnolog\'ias de la Informaci\'on y Comunicaciones, Universitat Polit\`ecnica de
  Val\`encia}, city={Val\`encia, Val\`encia}, country={Spain}}

%% ABSTRACT — <= 250 words, concise/factual, no references.
\begin{abstract}
The dosing of intravenous fluids and vasopressors in sepsis is a sequential
decision made under uncertainty and guided largely by clinical judgment, which
makes it a natural target for reinforcement learning from historical care.
Because a learned policy cannot be trialed on patients, its value must be
estimated off-policy, and such estimates can be fragile and optimistic. This work advances the reliable evaluation of sepsis treatment policies by combining off-policy estimation, reliability diagnostics, and clinician-agreement analyses in a transparent validation framework. We modeled fluid and vasopressor dosing on a cohort of
\num{36872} septic ICU stays drawn from the MIMIC-IV critical-care database, as a
discretized Markov decision process with \num{1000} states and \num{25} actions, defined by a five-by-five grid of fluid and vasopressor levels and solved by policy iteration. The clinicians' behavior policy
was estimated with a random forest, which mitigated the collapse of the Effective
Sample Size (ESS \num{50.1} against \num{4.0} with smoothed counts) that
otherwise destabilizes the importance-sampling estimate. The learned policy
was evaluated with two estimators, weighted importance sampling (WIS) and
fitted Q evaluation (FQE), with the ESS and clinician agreement as reliability
checks. An empirical variable selection found that the
composition of the state matters more than its size. Both estimators place the
learned policy above the clinicians' return (WIS \num{50.8} and FQE \num{46.8}
against \num{38.2}, ESS \num{50.1}), yet it departs only modestly from observed
practice (total variation \num{0.18}), favoring less intravenous fluid. These retrospective single-center off-policy results support the learned policy as a clinically plausible refinement of observed practice and motivate its further evaluation as a discordance-based clinical decision-support approach.
\end{abstract}

%% KEYWORDS — 1 to 7, English, avoid multi-word "X of Y".

\end{frontmatter}

%\linenumbers

%% ---------------------------------------------------------------------
%% 1. INTRODUCTION
%% ---------------------------------------------------------------------
\section{Introduction}
% Source: Cap. 01 Introducción.
Sepsis is defined by the Sepsis-3 consensus as life-threatening organ dysfunction caused by a dysregulated host response to infection \cite{singer2016}, and is one of the leading causes of death worldwide: the Surviving Sepsis Campaign estimates on the order of \num{49} million cases and \num{13} million related deaths each year \cite{prescott2026}, while survivors face a lasting burden of physical, cognitive and mental impairment known as post-sepsis syndrome \cite{prescott2018}. This definition is deliberately broad, so that two patients sharing the diagnosis can differ substantially in infection source, physiological course and treatment needs, and no single treatment pattern is optimal for all.
International guidelines set out the pillars of management, antimicrobials, source
control, intravenous fluid resuscitation and vasopressor support
\cite{prescott2026}, yet the dose of the two hemodynamic levers, fluids and
vasopressors, remains uncertain and guided largely by clinical judgment. Unlike antimicrobial
choice or source control, which are multi-step decisions difficult to encode from
structured data, fluids and vasopressors are administered as numeric doses at
defined times, which makes them well suited to a data-driven approach.

Dosing these levers is a sequential decision problem under uncertainty: the
clinician acts repeatedly over the stay, each action changing the patient's state
and shaping the next. Framed this way, with a state built from the patient's
physiology and an action given by the fluid and vasopressor doses, it is naturally
a Markov decision process (MDP), and seeking the dosing policy that maximizes estimated survival is
a reinforcement learning problem. The Markov assumption, that the current state
summarizes the relevant past, is an approximation to a problem that is strictly
partially observable, but it is the standard hypothesis in this line of work
\cite{komorowski2018}. Because a learned policy cannot be evaluated directly on patients without prior validation, learning and evaluation must first be offline, from the historical trajectories of the clinicians' own
policy. We work on MIMIC-IV \cite{johnson2023,mimiciv2024}, the deidentified
critical-care record of the Beth Israel Deaconess Medical Center (BIDMC) between
2008 and 2022.

The AI Clinician \cite{komorowski2018} established this framing for sepsis, learning
a fluid and vasopressor policy on the earlier MIMIC-III database
\cite{johnson2016} and reporting lower mortality when the clinicians' doses matched
its recommendation. Its central difficulty, shared by offline reinforcement learning
in health more broadly, is evaluation: a policy that is never executed can only be
assessed off-policy, from retrospective data, and such estimates are notoriously
fragile and optimistic \cite{gottesman2019}. A policy that departs far from observed
practice may in principle be more valuable, yet the same divergence leaves little
data to corroborate it, and a single importance-sampling estimate can be dominated
by a handful of trajectories. The gap we address is therefore the need for a disciplined off-policy evaluation of sepsis treatment policies, with reliability diagnostics that go beyond a single optimistic estimate, carried out on an updated cohort.
We revisit the problem on MIMIC-IV, a larger and structurally different cohort from
MIMIC-III, and frame the learned policy as a clinical plausible refinement of observed practice rather than a new treatment strategy. This interpretation suggests a discordance-based role for clinical decision support, in which the learned policy highlights well-supported departures from observed dosing practice rather than replacing clinical judgment.

The contributions of this work are the following:
\begin{itemize}
  \item An update of the AI Clinician's pipeline to MIMIC-IV, with a rigorous, empirical variable selection, which
  finds that the composition of the state matters more than its size and which
  removes the leakage and ad hoc patches of the original pipeline: every fitted
  quantity is estimated on the training split alone, fluids are not up-weighted and
  the vasopressor receives no bespoke transform.
  \item A dual off-policy evaluation pairing weighted importance sampling (WIS) and
  fitted Q evaluation (FQE), estimators whose failure modes differ, with the
  effective sample size (ESS) as a reliability diagnostic and clinician agreement as
  an independent check, in place of the single estimator of the reference work.
  \item A behavior-policy estimator based on a random forest, which controls the
  collapse of the effective sample size (\num{50.1} against \num{4.0} with smoothed empirical counts)
  that otherwise destabilizes the importance-sampling estimate.
  \item An honest analysis of how the effective sample size, WIS and FQE trade off
  against the number of states $K$, and of why $K=\num{1000}$ is a compromise rather
  than a clean optimum.
\end{itemize}

Under this evaluation, the learned policy remains close to observed clinical practice while being consistently favored by both off-policy estimators. We therefore interpret it as a clinically plausible refinement of care, whose value should be read through the reliability diagnostics developed throughout the paper.

%% ---------------------------------------------------------------------
%% 2. BACKGROUND / RELATED WORK
%% ---------------------------------------------------------------------
\section{Background and related work}
\label{sec:background}
% Source: Cap. 02 Estado del arte.

\subsection{Sepsis and hemodynamic management}
Sepsis is defined by the Third International Consensus (Sepsis-3) as
life-threatening organ dysfunction caused by a dysregulated host response to
infection, made operational as an acute increase of at least two points in the
Sequential Organ Failure Assessment (SOFA) score \cite{singer2016}. The SOFA
score, introduced by Vincent et al.\ \cite{vincent1996} and adopted by Sepsis-3,
grades six organ systems (respiratory, coagulation, hepatic, cardiovascular,
neurologic and renal) from \num{0} to \num{4} each, for a composite from \num{0}
to \num{24} in which higher values denote worse dysfunction; the criteria we
apply to delimit the cohort are given in Section~\ref{sec:data}. Hemodynamic
management rests on a few strong recommendations of the Surviving Sepsis
Campaign: crystalloids as the first-line resuscitation fluid, norepinephrine as
the initial vasopressor, and a mean arterial pressure target of \num{65} mmHg
\cite{prescott2026}. Beyond these pillars much of the secondary guidance is
conditional and rests on low-certainty evidence, and the concrete dose of fluids and vasopressors and their timing remain guided by clinical judgment. It is precisely this margin, which dose and when, that admits a data-driven approach.

\subsection{Reinforcement learning for sepsis treatment}
The AI Clinician \cite{komorowski2018} established the framing we adopt: it cast
fluid and vasopressor dosing as a sequence of decisions over the stay, built a
Markov decision process (Section~\ref{sec:mdp}) and solved it by policy iteration
to obtain a policy maximizing estimated survival. On its data that policy
attained a higher estimated value than the clinicians, and lower mortality was observed when the administered dose was closer to the recommended one, with a characteristic pattern of recommending less intravenous fluid and more low-dose vasopressor; these figures, however, come from a retrospective off-policy evaluation
and not from a prospective clinical trial. Its instantiation selected \num{48}
clinical variables, discretized each stay into \qty{4}{\hour} windows, clustered
the state space into $K=\num{750}$ discrete states, and defined \num{25} actions
as the $5\times5$ combinations of fluid and vasopressor dose (four nonzero
quartile levels plus zero); the reward was terminal, $+100$ for survival and
$-100$ for death at \num{90} days, with a discount $\gamma=\num{0.99}$. The model
was developed on MIMIC-III \cite{johnson2016} and validated externally on the
eICU Research Institute database. Subsequent work explored deep reinforcement
learning for the same task \cite{raghu2017}, a direction we return to only as
future work. We revisit the problem on MIMIC-IV \cite{johnson2023}, with the
departures from that reference design developed in the sections that follow: an
empirical variable selection (Section~\ref{sec:variable-selection}), a
random-forest behavior policy (Section~\ref{sec:behaviour-policy}) and a dual
off-policy evaluation (Section~\ref{sec:ope}).

\subsection{Off-policy evaluation and its pitfalls}
Because the learned policy is not executed in the observational data, it must first be judged off-policy, from trajectories generated by the clinicians. The clinicians' own return is
estimated directly, on-policy, from the observed outcomes \cite{sutton2020}; the
learned policy, by contrast, requires estimators that correct for the mismatch
between the two policies. Two complementary families are used in this literature,
and their derivations are deferred to Section~\ref{sec:ope}. Importance sampling,
and its self-normalized variant weighted importance sampling (WIS), reweights
each observed trajectory by how probable it would have been under the target
policy; the estimator is consistent but its variance can be enormous, in
principle unbounded \cite{ionides2008}, when a few trajectories carry
disproportionate weight, and its reliability is summarized by the effective
sample size \cite{martino2017}. Fitted Q-evaluation (FQE) \cite{le2019} instead
estimates the value of the target policy without importance ratios, avoiding that
variance at the cost of bias when the fitted model approximates the Bellman
operator poorly; the two therefore fail in complementary ways, variance against
bias.

These fragilities are not incidental. Gottesman et al.\ \cite{gottesman2019}, in methodological guidance for off-policy reinforcement learning in health, set out why
off-policy reinforcement learning from observational health data is hard to
trust: omitted state variables can confound the learned associations, an
instance of the Markov assumption failing; a policy is valuable precisely when
it departs from observed practice, yet the same departure leaves few
corroborating trajectories, so the more valuable policies are the harder ones
to evaluate; and a policy validated retrospectively need not transfer
prospectively, both because of
distribution shift across sites and time \cite{moreno-torres2012} and because a
sparse terminal reward may be too coarse a proxy for benefit. Raised from within
the field rather than against it, these concerns are the direct motivation for
the methodological emphasis of this work: a dual evaluation with explicit
reliability diagnostics in place of a single optimistic estimate.

%% ---------------------------------------------------------------------
%% 3. DATA AND COHORT
%% ---------------------------------------------------------------------
\section{Data and cohort}
\label{sec:data}
% Source: Cap. 03 Datos y cohorte.
We built the cohort on MIMIC-IV v3.1 \cite{johnson2023,mimiciv2024}, a public,
deidentified critical-care database drawn from the electronic health record of
BIDMC between 2008 and 2022. This updates the data source used by the AI Clinician \cite{komorowski2018}, which was developed on the now superseded MIMIC-III \cite{johnson2016} and validated externally on the eICU Research Institute database (eRI). MIMIC-IV is the current standard and roughly \num{1.5}
times larger (\num{94458} ICU stays against the \num{61532} of MIMIC-III), with a
broader catalog of derived clinical concepts. The dataset was obtained in
accordance with the guidelines set forth by the Massachusetts Institute of
Technology and the BIDMC Institutional Review Board. Access to the dataset was
granted upon completion of the required Collaborative Institutional Training
Initiative program course on data use and privacy for researchers
\cite{pollard2026}.

\subsection{Cohort selection}
\label{sec:cohort}
Sepsis was identified with the official \texttt{sepsis3} derived view, which
operationalizes the Sepsis-3 criteria: a suspected infection (an antibiotic order
paired with a culture sample within the prescribed window) together with an
increase of at least two SOFA points attributable to the episode. Sepsis onset,
the earlier of those two events, anchors the origin of every trajectory, and the
unit of analysis is the ICU stay. Beyond the Sepsis-3 definition we applied three
exclusions. Restricting to adults (age $\geq 18$) removed no one, since the
derived view already excludes pediatric ICUs, but we keep it explicit to fix the
adult scope. Two further filters removed stays that would distort the learning of
a dosing policy: treatment withdrawal, defined operationally as death in the last
\qty{24}{\hour} of the window in a patient who had received vasopressors that were
already stopped at the window's close, a pattern compatible with an end-of-life
decision rather than a therapeutic one; and the absence of any documented
intravenous fluid, which leaves no hemodynamic intervention to observe. Of the
\num{94458} ICU stays in MIMIC-IV, the adult Sepsis-3 definition delimited
\num{41295}; the withdrawal filter then discarded \qty{1.4}{\percent}
(\num{573} stays) and the no-fluid filter a further \qty{9.5}{\percent}
(\num{3850}), leaving a final cohort of \num{36872} ICU stays
(\SuppFigConsort).

These stays correspond to \num{28605} unique patients, the \num{8267} additional
stays being ICU readmissions of the same patient. The training and validation
partition (\num{80}/\num{20}) was therefore drawn by patient rather than by stay,
so that a patient's trajectories cannot be split across the two sets and leak
information. Data were extracted over the interval
$[\text{onset}-\qty{24}{\hour},\ \text{onset}+\qty{48}{\hour}]$, while the
effective MDP trajectory spans
$[\text{onset},\ \text{onset}+\qty{48}{\hour}]$ (Section~\ref{sec:preprocessing}).

\subsection{Cohort characteristics}
\label{sec:cohort-char}
The cohort is an adult critical-care population: median age \num{66} years
(interquartile range \numrange{55}{76}) and \qty{58.0}{\percent} male. Despite
roughly doubling the size of the original cohort (\num{36872} against \num{17083}
stays, from the larger MIMIC-IV), the demographic composition is almost identical
to the AI Clinician's (mean age \num{64.4} $\pm$ \num{16.9} years,
\qty{56.2}{\percent} male \cite{komorowski2018}), which indicates broad demographic similarity and leaves mortality as a relevant contrast (Section~\ref{sec:mortality}).

Baseline severity is likewise comparable: the mean SOFA at onset is \num{7.7} $\pm$ \num{2.6}, against \num{7.2} $\pm$ \num{3.2} in the reference study \cite{komorowski2018}. This figure requires a caveat. The MIMIC-IV derived view assigns zero to any organ system without a measurement in its window, which can underestimate SOFA when sampling is sparse, most acutely at onset. We therefore recompute SOFA on the imputed data, following the same official thresholds, and use this corrected score throughout: both for this descriptive comparison and as the state variable observed by the policy at every step. This keeps severity measurement consistent across descriptive reporting and policy learning, rather than correcting it only where it is reported. The residual approximations of this recomputation, and the sense in which cohort onset is still detected on the uncorrected score, are addressed in Section~\ref{sec:limitations}. Chronic disease burden, by the van Walraven adaptation
of the Elixhauser index \cite{elixhauser1998,vanwalraven2009}, has a mean of
\num{14.6} $\pm$ \num{10.0}.

The clinical state at onset is that of incipient organ dysfunction. Median vital
signs sit close to normal (heart rate \qty{86}{\bpm}, mean arterial pressure
\qty{77}{\mmHg}, oxygen saturation \qty{98}{\percent}, Glasgow Coma Scale 15), while
perfusion and organ-damage markers are already deranged in a relevant fraction of
patients: lactate is elevated at the median (\qty{2.1}{\milli\mole\per\liter}, third quartile
\num{3.3}) and creatinine reaches \qty{1.9}{\milli\gram\per\deci\liter} at the third quartile. The full
per-variable baseline is reported in the \SuppTabBaseline. In line with sex-reporting guidance we report the
sex distribution above; a sex-stratified analysis of the policy was not performed
and is noted as a generalizability limitation (Section~\ref{sec:limitations}).

\subsection{Mortality}
\label{sec:mortality}
MIMIC-IV provides three death indicators with different coverage. The in-hospital
mortality flag (\texttt{hospital\_expire\_flag}), available for \qty{100}{\percent}
of admissions, defines the terminal reward of the MDP (Section~\ref{sec:mdp}); the
time of in-hospital death (\texttt{deathtime}), present only for in-hospital deaths,
truncates trajectories and applies the withdrawal criterion; and the date of death
(\texttt{dod}), censored one year after discharge, yields \num{90}-day mortality as
a secondary outcome. Table~\ref{tab:mortality} contrasts the four descriptive
mortalities of our cohort with those the reference study reports on its development
(MIMIC-III) and external validation (eRI) cohorts. Mortality in our cohort is between \num{1.4} and \num{1.8} times higher than in the reference MIMIC-III cohort across all four indicators, but matches the external eRI cohort on the two comparable metrics
(in-hospital \qty{16.2}{\percent} against \qty{16.4}{\percent}, almost exactly;
ICU \qty{10.7}{\percent} against \qty{9.8}{\percent}, more loosely).
The absolute mortalities are therefore not directly comparable with Komorowski's MIMIC-III development cohort, but their agreement with the external eRI cohort supports the clinical plausibility of the case mix; this comparison is developed further in Section~\ref{sec:clinical-validity}.

\begin{table}[t]
\centering
\caption{Descriptive mortality of the MIMIC-IV cohort against the development
(MIMIC-III) and external validation (eRI) cohorts of the AI Clinician
\cite{komorowski2018}. In-hospital mortality (\texttt{hospital\_expire\_flag})
governs the terminal reward of the MDP; the remaining indicators are descriptive.
All values are percentages.}
\label{tab:mortality}
\begin{tabular}{lccc}
\toprule
Indicator & MIMIC-IV (ours) & MIMIC-III & eRI \\
\midrule
ICU mortality         & 10.7 & 7.4  & 9.8  \\
In-hospital mortality & 16.2 & 8.9  & 16.4 \\
28-day mortality      & 19.8 & 11.3 & n/a  \\
90-day mortality      & 27.3 & 18.9 & n/a  \\
\bottomrule
\end{tabular}\\[3pt]
{\footnotesize ICU mortality is death before ICU discharge; day-28 and day-90
mortality are counted from sepsis onset. Day-28 and day-90 mortality were not
available for the eRI cohort in the reference study.}
\end{table}

%% ---------------------------------------------------------------------
%% 4. METHODS
%% ---------------------------------------------------------------------
\section{Methods}
% Source: Cap. 04 Metodología.  CORE OF THE PAPER — draft first.

\subsection{Preprocessing pipeline}
\label{sec:preprocessing}
The state representation was designed to cover the six organ systems of the
SOFA score (respiratory, coagulation, hepatic, cardiovascular, neurologic and
renal), together with vital signs, lactate as a perfusion marker, and slowly
varying context (weight, mechanical ventilation, comorbidity by the van
Walraven adaptation of the Elixhauser index \cite{elixhauser1998,vanwalraven2009},
age and sex). Variables were extracted from the \emph{hosp} and \emph{icu}
modules of MIMIC-IV \cite{johnson2023} and from the \texttt{mimiciv\_derived}
concepts, following the AI Clinician \cite{komorowski2018} with the deliberate
departures detailed below. The pipeline is summarized in
\SuppFigPipeline; \SuppTabStateVariables  lists the state
variables of the final configuration, whose empirical selection is deferred to
Section~\ref{sec:variable-selection}.

Patient records were discretized into non-overlapping \qty{4}{\hour} windows
from sepsis onset. Within a window, measurements were aggregated by type: the
mean for vital signs and laboratory values, the sum for administered fluids and
urine output, the last value for the SOFA score and its components (so that a
transient spike does not inflate the score), the minimum for the Glasgow Coma
Scale (the worst neurologic state of the block), and the maximum for
vasopressors and mechanical ventilation. Interval-valued items (infusions,
ventilation, weight) were assigned to every window they overlap before
aggregation, so that no dose was split or double counted.

Missing values were imputed in two stages. First, a sample-and-hold
carry-forward with type-specific limits: \qty{8}{\hour} for vital signs,
\qty{12}{\hour} for blood gases, and \qty{24}{\hour} for laboratory values and
clinical scores; static or slowly varying variables were propagated without
limit in both directions. Continuous variables were then winsorized to their
0.1st and 99.9th percentiles and the remaining gaps filled with a
$k$-nearest-neighbours imputer ($k=5$). For any variable that was never measured
in at least \qty{15}{\percent} of patients, a binary \texttt{was\_measured}
indicator was added, so that a purely extrapolated value remains
distinguishable from an observed one.

Skewed continuous variables were transformed by $\log(1+x)$, which admits the legitimate zeros of urine output and vasopressor dose, and all continuous variables were then standardized to zero mean and unit variance; binary indicators were centered by their training prevalence without rescaling. Every fitted quantity (winsorizing thresholds, imputation neighbours, the scaler and the clustering below) was estimated on the training split only and applied unchanged to validation, avoiding leakage from validation into preprocessing. We also omitted two implementation-specific transformations used in the published AI Clinician pipeline: fluids were not up-weighted by a factor of two, and the vasopressor was not given the bespoke logarithm that implementation applies to it alone. Both interventions instead follow the same $\log(1+x)$ transformation used for every other skewed continuous variable (\SuppTabStateVariables).

\subsection{Markov decision process}
\label{sec:mdp}
We frame hemodynamic management as a Markov decision process, the tuple
$\langle \mathcal{S}, \mathcal{A}, T, R, \gamma \rangle$, estimated on the
training split.

\paragraph{States}
Each \qty{4}{\hour} window, represented by its scaled feature vector, was
assigned to the nearest of $K$ clusters by MiniBatch $k$-means
\cite{sculley2010,pedregosa2018}, chosen over classical $k$-means\texttt{++}
\cite{arthur2007} for the speed that made the hyperparameter sweep tractable.
The clustering was fit on the training split with a fixed seed and multiple
initializations, and outcome columns were excluded to prevent trivial leakage.
Two absorbing terminal states, hospital discharge and death, were appended,
giving $K+2$ states. The value of $K$ is a compromise: too large yields tiny
clusters with insufficient support, too small merges clinically distinct
situations; its final value $K=\num{1000}$ is justified in
Section~\ref{sec:configuration}.

\paragraph{Actions}
The two therapeutic levers, intravenous fluid volume and vasopressor dose, were
each discretized into five levels, giving a $5\times5=25$ action grid
\cite{komorowski2018}. Within a window, fluids were summed and the vasopressor
was taken as the maximum norepinephrine-equivalent dose. Level~0 is no
administration; levels~1 to 4 are delimited by the 25th, 50th and 75th
percentiles of strictly positive doses on the training split, with zeros
excluded, since the large fraction of untreated windows would otherwise collapse
the quartiles toward the origin. \SuppTabActions  reports the resulting
cutoffs. The action $(0,0)$, no fluid and no vasopressor administration, serves as the reference when comparing policies with clinician behavior.

\paragraph{Transitions and reward}
Transition probabilities $T(s,a,s')$ were estimated as the relative frequencies
of the observed next state, with each patient's last window transitioning to the
absorbing state of its outcome. Triplets $(s,a,s')$ observed fewer than five times were excluded from the transition estimate as insufficiently supported, which leaves some state-action pairs with an empty transition row; these pairs are subsequently masked during policy improvement (Section~\ref{sec:policy-iteration}). Rewards are terminal and sparse: $+100$ on
survival to hospital discharge and $-100$ on in-hospital death (from
\texttt{hospital\_expire\_flag}), with zero intermediate reward. With zero intermediate reward, the expected immediate reward for a state-action pair is determined by the probability of transitioning to either absorbing terminal state,
\begin{equation}
R(s,a) = \Pr(s'=\mathrm{discharge}\mid s,a)\,(+100)
       + \Pr(s'=\mathrm{death}\mid s,a)\,(-100).
\label{eq:reward}
\end{equation}
Trajectories were truncated at
$\min(\mathrm{deathtime},\ \mathrm{ICU\ discharge},\ \mathrm{onset}+\qty{48}{\hour})$.

An optional penalty on the immediate reward lets the search express a
preference for conservative dosing. Writing an action as its pair of fluid and
vasopressor levels, each on the 0 to 4 scale of the action grid, the penalized
reward is
\begin{equation}
R_P(s,a) = R(s,a) - \lvert P\rvert\,
  \bigl(\mathrm{level}_{\mathrm{fluid}} + \mathrm{level}_{\mathrm{vaso}}\bigr),
\label{eq:reward-penalty}
\end{equation}
so that no intervention $(0,0)$ is unpenalized and the maximal dose on both
levers is penalized most, by $8\lvert P\rvert$, while preserving the terminal $\pm100$ scale.
Both the penalty strength $P$ and the discount $\gamma$ were explored in the
sweep; their final values, $P=0.05$ and $\gamma=0.95$, are set out and justified
together with the rest of the final configuration in
Section~\ref{sec:configuration}.

\subsection{Policy iteration and action masking}
\label{sec:policy-iteration}
The MDP was solved by policy iteration \cite{sutton2020}, alternating policy
evaluation and improvement from a random initial policy. Evaluation solves the
Bellman equation for the current policy $\pi$,
\begin{equation}
V_\pi(s) = R(s,\pi(s)) + \gamma \sum_{s'} T(s,\pi(s),s')\,V_\pi(s'),
\label{eq:bellman}
\end{equation}
iterating until the largest change in $V_\pi$ falls below $10^{-6}$; improvement
then recomputes, for every action,
\begin{equation}
Q(s,a) = R(s,a) + \gamma \sum_{s'} T(s,a,s')\,V_\pi(s'),
\label{eq:qvalue}
\end{equation}
and updates $\pi$ toward $\arg\max_a Q(s,a)$ until the policy is stable.

A plain argmax over $Q$ does not distinguish well-estimated actions from rarely
observed ones. A pair with little support carries an unreliable $R$ and $T$, and
in the limit of an empty transition row $Q(s,a)\approx 0$, which the argmax would
favor over well-estimated actions of negative value. We therefore mask
unreliable actions before the argmax, setting $Q(s,a)=-\infty$ whenever the pair
was observed fewer than $M=25$ times on the training split, or its transition
row was emptied by the support filter. The threshold $M$ is a compromise: too
high and the policy merely imitates the clinicians, too low and it rests on a
handful of cases; its value is reported in Section~\ref{sec:configuration}.

When every action in a state is masked, we fall back to the clinicians' modal
action there, a deliberately conservative choice that defers to observed
practice under absent evidence; the two absorbing states are excluded from this
computation. Policy iteration returns a deterministic policy, which we soften to
an $\varepsilon$-soft form ($\varepsilon=0.01$) for the off-policy evaluation
described next.

\subsection{Behavior policy estimation}
\label{sec:behaviour-policy}
Weighted importance sampling requires the clinicians' behavior policy $\pi_b$,
which is not observed: the records show the actions taken, not the
probabilities behind them. We therefore estimate $\pi_b(a\mid s)$ and form, at
each step, the importance ratio against the learned policy $\pi_e$,
\begin{equation}
\rho_t = \frac{\pi_e(a_t\mid s_t)}{\pi_b(a_t\mid s_t)}.
\label{eq:ratio}
\end{equation}

We model $\pi_b$ with a random forest \cite{breiman2001} trained on the
\emph{continuous} patient state rather than on the discrete cluster, with 100
trees, a maximum depth of 20, and a minimum of 20 samples per leaf, conservative
defaults for a forest of this size chosen to avoid overfitting the propensity
model; these values were fixed a priori and not included in the sweep, so that
$\pi_b$ itself would not be tuned against the same evaluators it feeds into. The predicted
probabilities were mixed with a uniform distribution (weight $\alpha=0.05$) so
that no observed clinician action receives zero probability and no ratio
diverges.

Estimating $\pi_b$ on the continuous state, while $\pi_e$, the transition and
reward models, and the FQE estimator below all operate on the discrete
clusters, is a deliberate asymmetry. Discretization is needed to keep the MDP
and policy iteration tractable, but $\pi_b$ enters only as the denominator of
Equation~\eqref{eq:ratio}, where its errors are amplified by the division and
matter most \cite{raghu2018}. An empirical count assigns a single distribution
to every patient in a cluster, whereas the forest captures within-cluster
variation, giving a more faithful model of that denominator without
compromising the MDP.

We validated this choice with a control experiment that reuses one fixed
trained model and swaps only the $\pi_b$ estimator, smoothed empirical counts
versus the random forest, so as to isolate its effect on the evaluation
(Section~\ref{sec:ope-results}). A plausible mechanism, which we report as an argued
hypothesis rather than an instrumented fact, is that the large fraction of unobserved
state-action cells forces the smoothed count to a near-zero floor, so that a single
high-ratio step dominates the trajectory weight.

\subsection{Off-policy evaluation}
\label{sec:ope}
We estimate the value of the learned policy on the held-out validation split
with two complementary estimators whose failure modes differ, weighted
importance sampling (WIS) and fitted Q evaluation (FQE), together with the
effective sample size (ESS) as a reliability diagnostic and clinician agreement
as an independent check. Both estimators evaluate exactly the same policy: the
deterministic output of policy iteration is softened to an $\varepsilon$-soft
form that assigns $1-\varepsilon$ to the recommended action and spreads
$\varepsilon$ over the remaining 24 actions, with $\varepsilon=0.01$.

\paragraph{Weighted importance sampling}
For each validation trajectory $i$, the per-step ratios of
Equation~\eqref{eq:ratio}, evaluated on the action the clinician actually took,
are multiplied into a cumulative weight; each ratio is clipped to a maximum
$C=20$, chosen empirically so that clipping affects a negligible fraction of
steps while still bounding the weight explosion at the cost of a controlled
bias \cite{ionides2008}, before multiplication,
\begin{equation}
w_i = \prod_{t} \min(\rho_{i,t},\,C).
\label{eq:weight}
\end{equation}
Writing $G_i=\gamma^{L_i}(\pm 100)$ for the discounted terminal return of a
trajectory of length $L_i$, the estimator is the self-normalized weighted mean
\begin{equation}
\hat V_{\mathrm{WIS}} = \frac{\sum_i w_i\,G_i}{\sum_i w_i}.
\label{eq:wis}
\end{equation}
Confidence intervals were obtained by bootstrapping patients with replacement
(2000 resamples; 2.5th and 97.5th percentiles) \cite{komorowski2018}; the same
procedure applied to the observed returns yields the clinicians' empirical
value, used as the reference.

\paragraph{Effective sample size}
The reliability of Equation~\eqref{eq:wis} was summarized by
\begin{equation}
\mathrm{ESS} = \frac{\left(\sum_i w_i\right)^2}{\sum_i w_i^2},
\label{eq:ess}
\end{equation}
which falls when a few trajectories carry most of the weight and so foreshadows
wide intervals \cite{martino2017}; we also report the fraction of clipped steps.

\paragraph{Fitted Q evaluation}
FQE takes the opposite approach: rather than reweighting observed trajectories,
it estimates the value of $\pi_e$ directly on the estimated environment, without
importance ratios \cite{le2019}. We iterate the Bellman evaluation of
Equation~\eqref{eq:bellman} for $\pi_e$ on $\hat T$ and $\hat R$, from
$V\equiv 0$ to the same $10^{-6}$ tolerance, and average the resulting value
over the patients' observed initial states,
\begin{equation}
\hat V_{\mathrm{FQE}} = \frac{1}{n}\sum_{i} \hat V^{\pi_e}\!\left(s_{i,0}\right).
\label{eq:fqe}
\end{equation}
In the tabular case this model-based value coincides with FQE. The two
estimators fail differently: WIS inflates in variance when the policies
diverge, whereas FQE is biased if the estimated dynamics are wrong, so their
agreement raises confidence and their disagreement flags a problem.

\paragraph{Model selection}
Configurations were compared by a rule fixed in advance, in three successive
gates: a reliability gate discarding any configuration whose ESS falls below a
preset floor of 50; a clinical-plausibility gate favoring, among the remaining configurations, policies whose non-intervention rate stays close to the
clinicians'; and a triangulation criterion preferring, among those, policies
that both WIS and FQE place above the clinicians' return. There is no universally agreed cutoff for this
quantity: 50 is a pragmatic minimum, chosen over a markedly laxer alternative
near ESS $\approx$ 4, since too low an effective sample size no longer supports
the trajectories with an adequate sample. Clinician agreement, the fraction of steps where
the recommended and observed actions coincide over the 25-action grid,
accompanies the reading but does not decide it. The search ranged over the
state-variable set, the number of states $K$, the support threshold $M$, the
discount $\gamma$, and the dose penalty $P$; the explored values are reported in
Section~\ref{sec:configuration}.

A fourth robustness layer was added after the three prespecified gates, which left
several dozen configuration survivors close enough in point margin that ranking by the point
estimate alone proved unstable: some configurations with a better margin than
the eventual winner turned out to carry a confidence interval for WIS that
barely, or did not, clear the clinicians' return. We therefore added a
robustness check, applied identically to every configuration surviving the first three
gates rather than singled out afterward, requiring the \emph{lower} bound of
the bootstrap interval, not just the point estimate, of both WIS and FQE to
exceed the clinicians' return. This extends the triangulation already in place
from a comparison of means to a comparison of intervals, the same disciplined
triangulation applied one level stricter, and its effect on the final choice is
reported in Section~\ref{sec:configuration}.

%% ---------------------------------------------------------------------
%% 5. RESULTS
%% ---------------------------------------------------------------------
\section{Results}
% Source: Cap. 05 Resultados.
We first report the two experiments that fixed the final configuration, the
variable set (Section~\ref{sec:variable-selection}) and the hyperparameter sweep
(Section~\ref{sec:configuration}), and then characterize the selected model: its
learned policy against the clinicians (Section~\ref{sec:learned-policy}), its
dual off-policy evaluation (Section~\ref{sec:ope-results}), the clinician
agreement (Section~\ref{sec:agreement}), and the observed relation between dose
divergence and mortality (Section~\ref{sec:dose-mortality}). This section reports the observable results; their interpretation, including the tensions exposed by the sweep, is deferred to the Discussion. All
configurations were scored on the held-out validation split.

\subsection{Variable selection}
\label{sec:variable-selection}
An early observation motivated a dedicated variable-selection experiment: on the
full declared state set, every policy produced by the sweep assigned no
intervention (action $(0,0)$) to close to \qty{80}{\percent} of states, reaching
\qty{85}{\percent} in the configurations with the most states. These degenerate
policies maximized both estimators by ceasing to treat, which an audit traced to
a block of variables that were missing in most windows and imputed almost
entirely, contributing near-constant per-patient values rather than
within-stay signal.

Two experiments then separated the roles of variable composition and count.
Composition dominated: at equal size, randomly drawn sets reached an
intervention agreement near \qty{7}{\percent} against roughly \qty{17.5}{\percent}
for the curated set, and almost all of that margin was carried by a single
variable, the fluid volume administered in the window, which alone accounted for
\qty{56}{\percent} of the variance in the policy's intervention rate. That a
single treatment variable in the state explains so much of the intervention rate
illustrates the state-action circularity we return to as a
limitation (Section~\ref{sec:limitations}). The same experiment discarded the
\texttt{was\_measured} presence indicators: adding them did not improve
intervention agreement at any size, with a mean paired difference from $-0.17$ to
$-1.15$ points, so the final set operates on clinical variables only.

The experiment also showed that intervention agreement is almost perfectly
correlated with the intervention rate itself (coefficient 0.99), so rewarding it
would reward encoding the administered treatment in the state rather than
clinical quality; the plausibility gate therefore uses proximity to the
clinicians' non-intervention rate, not intervention agreement. Finally, the
best-performing curated set was not defensible as a final representation, since
the coverage filter had left it without a single laboratory variable. The final
set was therefore built deliberately, combining sufficient coverage with the
forced inclusion of clinically essential markers. Five candidate sets spanning a
minimal clinical core to the full pool (\SuppTabCandidateSets) were
carried into the final sweep.

\subsection{Final configuration and diagnostics}
\label{sec:configuration}
The final sweep crossed the five variable sets with the four tabular
hyperparameters, the number of states $K$ (400 to 1200), the support threshold
$M$ (10 to 25), the discount $\gamma$ (0.95, 0.99, 0.999) and the dose penalty
$P$ (0 to 0.15), for \num{1200} policies in total, with the SOFA component of
the state recomputed on the imputed data rather than taken zero-filled from the
derived view (Section~\ref{sec:cohort-char}), and applied the decision rule of
Section~\ref{sec:ope}.

The plausibility gate separated the sets at once: the full set and the Komorowski replica left the policy without
intervention in about two thirds of steps on average (\qty{68}{\percent} and
\qty{65}{\percent}), far from the clinicians' \qty{35}{\percent}, reproducing
the collapse that motivated the analysis, whereas the curated sets stayed near
the clinical margin (about \qty{37}{\percent} for the main set and
\qty{35}{\percent} for the minimal core). The reliability gate was far more
selective: of the \num{1200} policies, \num{77} reached an
effective sample size of at least 50, all of them at the lower discount
$\gamma=0.95$; not one configuration survived at $\gamma=0.99$ or
$\gamma=0.999$ (Section~\ref{sec:ope-tension}). The plausibility gate then discarded
\num{12} more configurations whose non-intervention rate departed from the
clinicians' by more than 10 percentage points, leaving \num{65}; requiring both
WIS and FQE to exceed the clinicians' return narrowed this to \num{26}
configuration survivors, all from the main variable set at $\gamma=0.95$.

Among those 26, ranking by point margin alone would favor several
configurations at $K=800$, $M=15$, but their advantage does not survive a
fourth, additional robustness check: requiring the \emph{lower} confidence bound of
both WIS and FQE, not just their point estimate, to clear the clinicians'
return. Under this stricter reading only \num{3} configurations pass, all at
$K=\num{1000}$, and they narrow the choice of $M$ and $P$. $M=20$, $P=0$ ranks
highest on point margin but its WIS interval barely clears the clinicians' (the
lower bound sits at $+0.29$ over them), and it would require re-justifying a
support threshold different from the rest of the sweep; the remaining two, both
at $M=25$, split on $P$: $P=0.1$ is dominated by $P=0.05$ on FQE (\num{46.0}
against \num{46.8}) and on model calibration (TD-error 95th percentile
\num{75.8} against \num{74.3}) with an identical WIS.
The selected configuration (Table~\ref{tab:final-config}) is therefore
$K=\num{1000}$, $M=25$, $\gamma=0.95$, $P=0.05$: not the closest to the
clinicians' margin, but the one whose advantage over them is most robust to
sampling uncertainty in both estimators at once. The dose penalty is not left at zero: among the three finalists, $M$ and $P$
are what still distinguish them, and $P=0.05$ gives the most comfortable
worst-case margin of the three. Under the recomputed score
$\gamma=0.99$ has no survivor at all, so $\gamma=0.95$ was retained because it was the only discount with reliable surviving configurations, not because it was expected a priori.

This selection is not a blind fit to the evaluators. The rule, an ESS floor, a
plausibility gate, then WIS/FQE triangulation, was fixed in
Section~\ref{sec:ope} before the sweep was run and applied identically to all
\num{1200} candidates, so no configuration was singled out after the fact; the
confidence-interval layer that resolves the final choice was applied with the
same discipline to all 26 configuration survivors (Section~\ref{sec:ope}). Within that rule, triangulation only
breaks ties among the handful of policies that already clear the first gates:
it is not free to select any value that merely maximizes an estimator. The same
caution about optimizing directly against WIS and FQE motivated the
variable-selection criterion of Section~\ref{sec:variable-selection}, where
maximizing either estimator was found to reward non-intervention rather than
clinical quality; here the estimators again only adjudicate among plausible,
reliable survivors, not search an unconstrained space.

Table~\ref{tab:final-config} also reports the sizing diagnostics of the trained
model. One sizing trade-off, taken up in the Discussion
(Section~\ref{sec:ope-tension}), is already visible here: the two return
estimators do not respond alike to $K$, and the effective sample size peaks at
a smaller $K$ than WIS does, so the three cannot be maximized together.

%\begin{table}[t]
%\centering
%\caption{Final configuration selected by the decision rule, with the sizing
%diagnostics of the trained model on the validation split. The 22 state variables
%are listed in Table~\ref{tab:state-variables}.}
%\label{tab:final-config}
%\begin{tabular}{ll}
%\toprule
%Component & Value \\
%\midrule
%\multicolumn{2}{l}{\emph{Configuration}} \\
%State-variable set & Main set (\texttt{core\_hi75}), 22 variables. \\
%Number of states $K$ & \num{1000} (plus 2 absorbing) \\
%Support threshold $M$ & 25 \\
%Discount $\gamma$ & 0.95 \\
%Dose penalty $P$ & 0.05 \\
%$\varepsilon$-soft smoothing & 0.01 \\
%Terminal reward & $+100$ discharge / $-100$ death \\
%Behavior policy $\pi_b$ & Random forest \\
%\midrule
%\multicolumn{2}{l}{\emph{Sizing diagnostics}} \\
%Active states in validation & 1000 of \num{1000} \\
%States falling back to clinicians' mode & \qty{3.1}{\percent} (31 of 1000) \\
%Small clusters & \qty{3.2}{\percent} (32) \\
%State-action pairs with empty transition row & \qty{81.7}{\percent} \\
%\bottomrule
%\end{tabular}
%\end{table}

\begin{table}[t] \centering \caption{Final configuration selected by the decision rule, with the sizing diagnostics of the trained model on the validation split. The 22 state variables are listed in \SuppTabStateVariables.} \label{tab:final-config} \small \begin{tabularx}{\linewidth}{lX} \toprule Component & Value \\ \midrule \multicolumn{2}{l}{\emph{Configuration}} \\ State-variable set & Main set (\texttt{core\_hi75}), 22~variables \\ Number of states $K$ & \num{1000} plus 2 absorbing states \\ Support threshold $M$ & 25 \\ Discount $\gamma$ & 0.95 \\ Dose penalty $P$ & 0.05 \\ $\varepsilon$-soft smoothing & 0.01 \\ Terminal reward & $+100$ hospital survival / $-100$~in-hospital death \\ Behavior policy $\pi_b$ & Random forest \\ \midrule \multicolumn{2}{l}{\emph{Sizing diagnostics}} \\ Active states in validation & 1000 of \num{1000} \\ States falling back to clinicians' mode & \qty{3.1}{\percent} (31 of 1000) \\ Small clusters & \qty{3.2}{\percent} (32) \\ State-action pairs with empty transition row & \qty{81.7}{\percent} \\ \bottomrule \end{tabularx} \end{table}

\subsection{Learned policy vs clinicians}
\label{sec:learned-policy}
\begin{figure}[t]
  \centering
  \includegraphics[width=\linewidth]{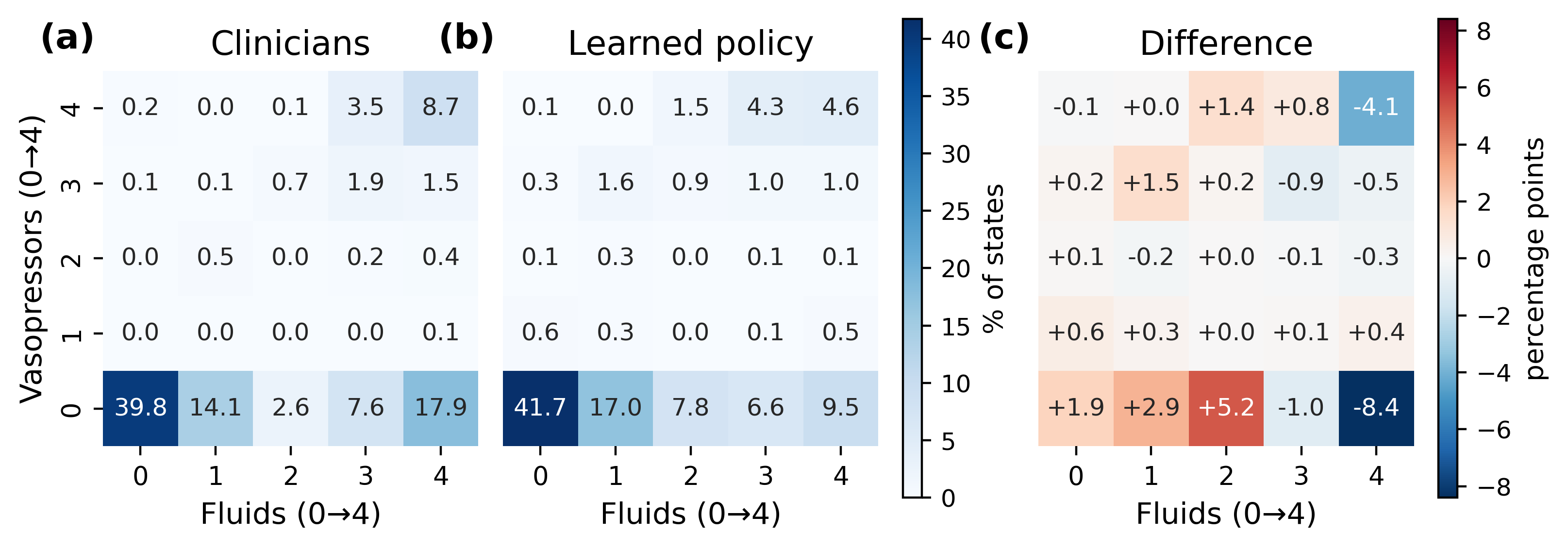}
  \caption{Modal action over the $5\times5$ grid of fluid (horizontal, 0 none to
  4 maximal) and vasopressor (vertical) levels, as the percentage of states.
  (a) Clinicians. (b) Learned policy. (c) Difference in percentage points,
  learned minus clinicians (red, actions the policy favors more than
  clinicians; blue, less).}
  \label{fig:policy-heatmaps}
\end{figure}

Before evaluating the policy we describe it. Figure~\ref{fig:policy-heatmaps}
shows, over the $5\times5$ grid of fluid and vasopressor levels, the modal
action per state for the clinicians (panel a) and for the learned policy
(panel b), together with their difference (panel c). The two policies share
the coarse structure, with the mass
concentrated in the vasopressor-free row and, within it, at the low fluid levels.
The learned policy shifts mass from maximal to intermediate fluids and leaves the
vasopressor axis almost unchanged, a direction consistent with the ``less
intravenous fluid'' tendency reported for the AI Clinician \cite{komorowski2018}.
Overall its action distribution differs from the clinicians' by a total variation
distance of \num{0.18}. Crucially the non-intervention collapse does not return: the
learned policy selects action $(0,0)$ in \qty{42.8}{\percent} of steps against the
clinicians' \qty{35.0}{\percent}, a gap of \num{7.8} percentage points rather than
the collapse to \qtyrange{80}{85}{\percent} that motivated the variable
selection.

\subsection{Off-policy evaluation}
\label{sec:ope-results}
\begin{figure}[t]
  \centering
  \includegraphics[width=\linewidth]{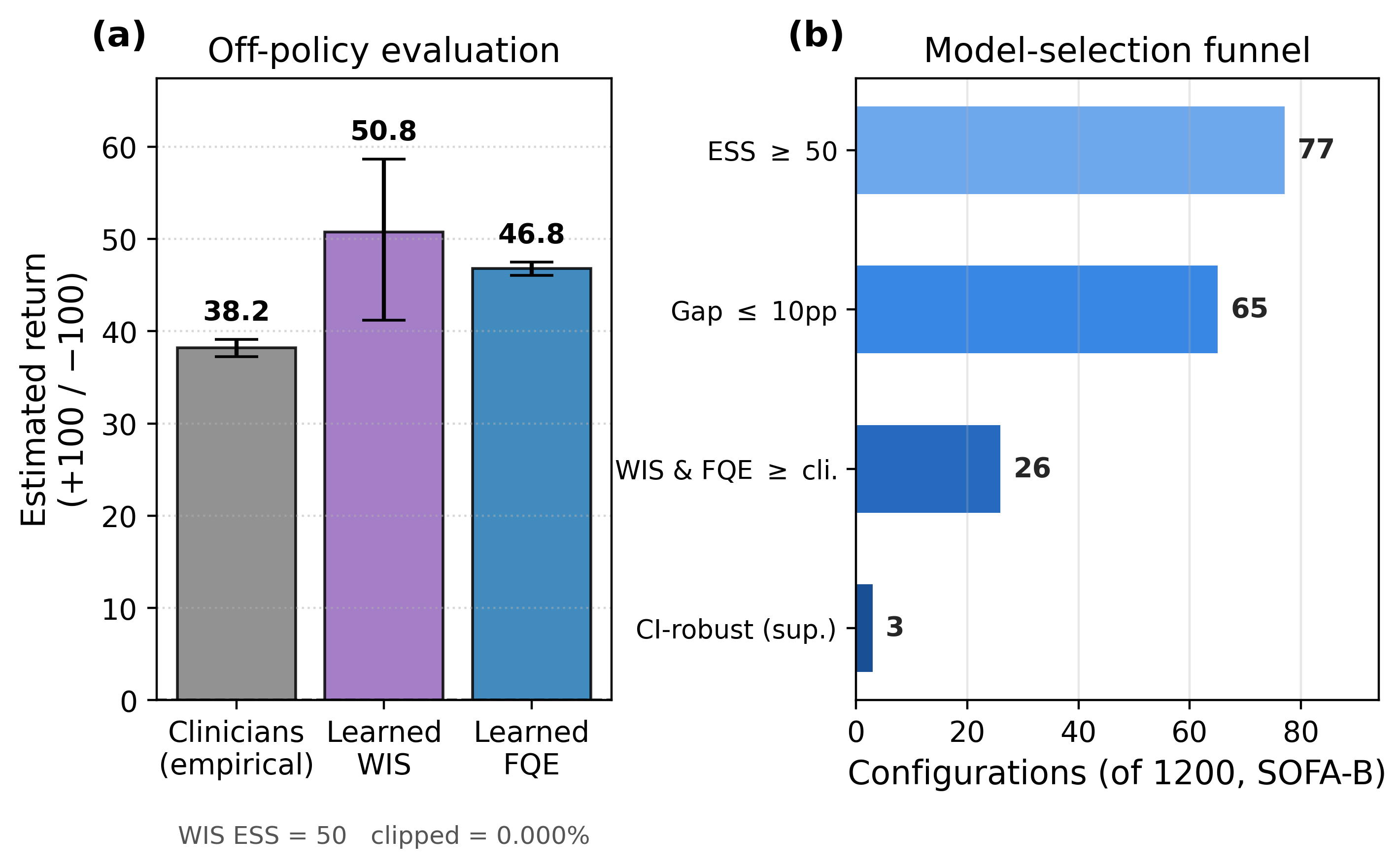}
  \caption{Off-policy evaluation of the final configuration. (a) Estimated return
  of the clinicians' policy (empirical) and of the learned policy by weighted
  importance sampling (WIS) and fitted Q evaluation (FQE), on the terminal-reward
  scale, with bootstrap \qty{95}{\percent} confidence intervals. (b) The
  decision-rule funnel applied to the \num{1200} candidates of the sweep: an
  ESS floor of 50 (\num{1200}$\to$\num{77}), a plausibility gate on the
  non-intervention rate (\num{77}$\to$\num{65}), triangulation requiring both
  WIS and FQE above the clinicians' return (\num{65}$\to$\num{26}), and the
  confidence-interval robustness layer of Section~\ref{sec:ope}
  (\num{26}$\to$\num{3}) that resolves the final choice among $K=\num{1000}$,
  $M=25$ survivors.}
  \label{fig:ope}
\end{figure}

On the final configuration the two estimators agree in placing the learned policy
above the clinicians (Figure~\ref{fig:ope}a). The clinicians' empirical return is
\num{38.2} [\num{37.2}, \num{39.1}] on the terminal-reward scale ($+100$
discharge, $-100$ death; the lower discount of the final configuration compresses
this scale relative to $\gamma=0.99$, so it is not comparable in magnitude to a
return reported at a different discount). Weighted importance sampling gives the
learned policy \num{50.8} [\num{41.2}, \num{58.6}], and fitted Q evaluation gives
\num{46.8} [\num{46.1}, \num{47.5}]; both clear the clinicians' return on the point estimate and on the lower bound of the interval. Both therefore satisfy the triangulation criterion, and both
survive the stricter confidence-interval check of Section~\ref{sec:ope}. The
estimate is reliable, though less comfortably than the point estimate alone
suggests: the WIS effective sample size is \num{50.1}, only just above the preset
floor of \num{50}, and the fraction of clipped steps is zero.

The reliability of WIS rests on the behavior-policy estimator. To isolate the effect of the behavior-policy estimator, we reused one fixed trained model, its clustering, MDP and policy $\pi_e$, and evaluated it twice, changing only the $\pi_b$ estimator. The metrics that depend
only on $\pi_e$, such as clinician agreement and FQE, were identical across the
two variants, and only the importance-sampling quantities moved. They moved
sharply: the effective sample size, \num{50.1} with the random forest, collapsed to
\num{4.0} with the smoothed empirical counts, so the estimate came to rest on a handful
of trajectories. WIS rose from \num{50.8} to \num{58.1}, but that increase cannot
be read as an improvement; it is the effect of a few extreme weights on an
already degraded estimate. With the random forest the effective sample size clears
the reliability floor and the intervals are narrower.

\subsection{Clinician agreement}
\label{sec:agreement}
Clinician agreement, the fraction of steps where the recommended and observed
actions coincide, was \qty{44.4}{\percent} overall and strongly asymmetric
(Figure~\ref{fig:agreement-dose}a). It was high on vasopressors
(\qty{79.0}{\percent}) and moderate on fluids (\qty{52.0}{\percent}), and rose to
\qty{69.4}{\percent} of steps when a one-level dose deviation is admitted as a
match. The sharpest contrast appears when conditioning on the clinician's
decision: on steps where the clinician did not intervene the policy agreed (also
not intervening) in \qty{94.4}{\percent}, whereas on steps where the clinician did
administer treatment it matched the exact action in only \qty{17.4}{\percent}.
These two figures are not measured on the same criterion, since non-intervention
is a single action while intervention requires matching the exact cell of the
dose grid; admitting a one-level tolerance, the intervention agreement rises to
\qty{53.8}{\percent}. By outcome, agreement was higher for survivors than for non-survivors
(\qty{46.1}{\percent} versus \qty{35.7}{\percent}, a difference of \num{10.4}
points). The reading of this asymmetry is deferred to the Discussion.

\subsection{Dose divergence and mortality}
\label{sec:dose-mortality}
\begin{figure}[t]
  \centering
  \includegraphics[width=\linewidth]{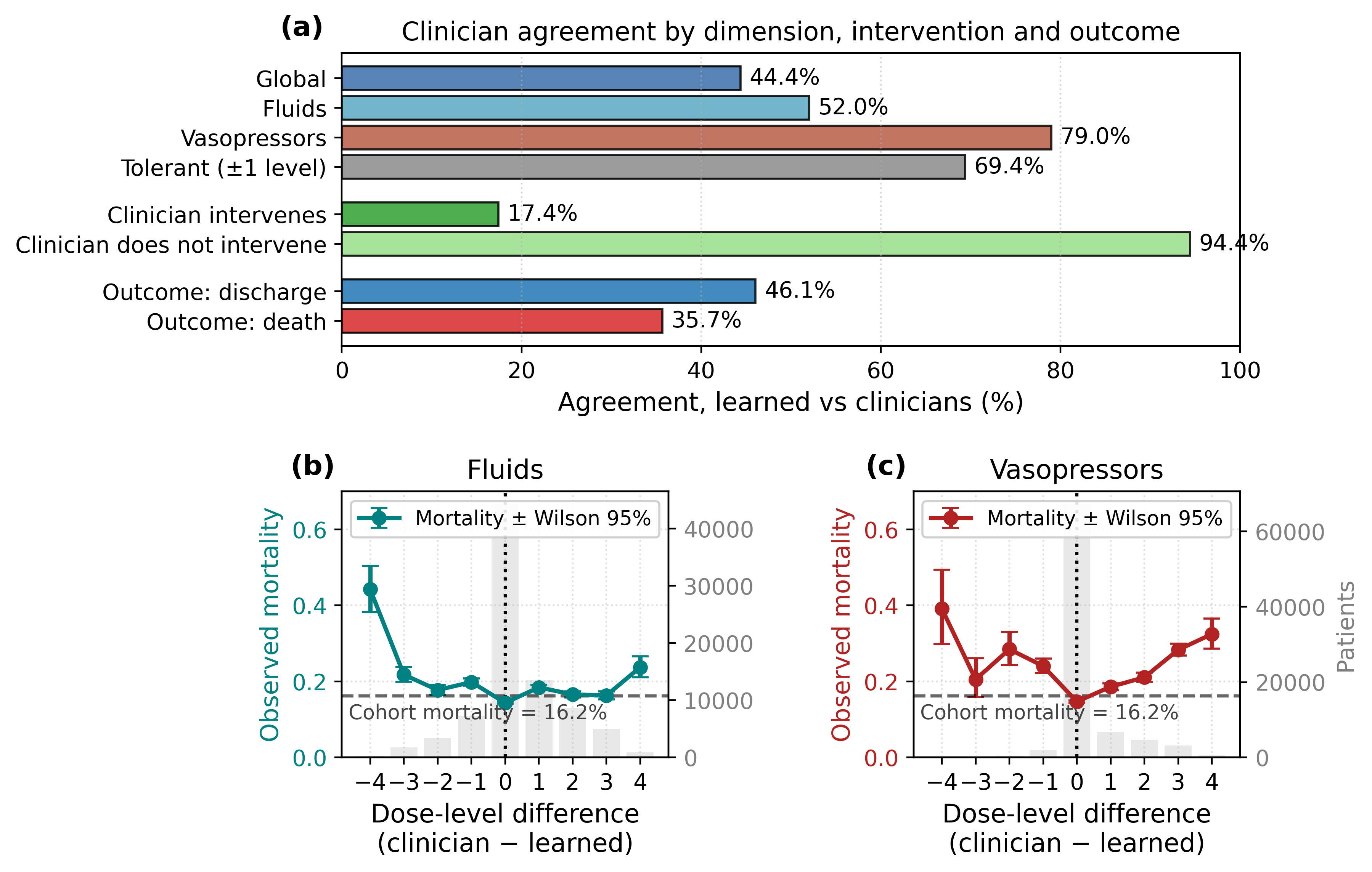}
  \caption{(a) Agreement between the learned policy and the clinicians, by
  treatment dimension (global, fluids, vasopressors and with a one-level
  tolerance), by whether the clinician intervenes, and by outcome (discharge
  versus death). (b, c) Observed in-hospital mortality on the validation split
  against the dose-level difference between clinician and policy (clinician minus
  policy; 0 is exact match) for fluids (b) and vasopressors (c); gray bars are
  patients per bin, error bars are \qty{95}{\percent} Wilson intervals, the
  horizontal dashed line is the cohort mortality and the vertical line the exact
  match.}
  \label{fig:agreement-dose}
\end{figure}

A final descriptive cut relates dose divergence to observed mortality. Validation
steps were grouped by how far the clinician's dose departs from the one the
learned policy recommends, and the observed in-hospital mortality was measured in
each group, for fluids and for vasopressors, following the scheme of the AI
Clinician \cite{komorowski2018} (Figure~\ref{fig:agreement-dose}b,c). In both
levers, observed mortality is lowest when the clinician's action coincides with
the recommendation and rises steadily as the two diverge in either direction.
These curves are an observed association, not causal evidence: the divergence may
itself reflect that the sickest patients receive the most extreme, and hence most
divergent, treatments. Their interpretation is taken up in the Discussion.

%% ---------------------------------------------------------------------
%% 6. DISCUSSION
%% ---------------------------------------------------------------------
\section{Discussion}
\label{sec:discussion}
% Source: Cap. 06 Discusión.

\subsection{Significance of this work}
The main contribution of this work is a more trustworthy account of a learned sepsis treatment policy, rather than an unqualified claim of a better policy. Compared with the original AI Clinician pipeline, this work replaces leakage-prone and implementation-specific preprocessing choices with an empirical variable-selection procedure and training-split-only estimation. Where a single optimistic estimator can be dominated by a handful of trajectories, the evaluation here pairs WIS and FQE, estimators with complementary failure modes, with an explicit reliability floor, the effective sample size, and an independent clinician-agreement check. It also reports how these quantities trade off as $K$ grows. The result of that scrutiny is deliberately modest: a policy that departs only slightly from observed practice and is best read as a clinically plausible refinement of care rather than as evidence of dramatic superiority. The significance of the work lies in that discipline: showing what a defensible off-policy evaluation of a clinical policy can look like, and being explicit about the limits of what it can and cannot claim.

\subsection{Interpreting the learned policy}
\label{sec:policy-interpretation}
The learned policy should be interpreted as a refinement of observed clinical practice rather than as a qualitatively new strategy: its action distribution departs from the clinicians' by a total variation of only \num{0.18} (Section~\ref{sec:learned-policy}). This is the defining tension of the setting. A more divergent policy might in principle hold greater clinical value, but retrospective off-policy evaluation would then have less data with which to support it. The compromise reached here is a policy close enough to observed practice to be empirically supported, which is also what makes a future decision-support role plausible (Section~\ref{sec:clinical-validity}).

Notably, that refinement avoids the non-intervention collapse that motivated the
variable selection: the policy withholds treatment in a clinically reasonable
\qty{42.8}{\percent} of steps (Section~\ref{sec:learned-policy}), far from the
\qtyrange{80}{85}{\percent} of the degenerate policies, rather than learning to stop
treating. Its characteristic recommendation, less intravenous fluid, coincides in
direction with the AI Clinician \cite{komorowski2018}, but the coincidence must be
read for what it is: each conclusion is a contrast against the clinicians of its own
cohort, and those baselines differ, since Komorowski contrasts with the MIMIC-III
clinicians and we with the MIMIC-IV clinicians, over different time windows, cohort
criteria and variable sets. This is a convergence of direction against the local
clinician, not of absolute dose, and it is confined to one lever: our policy barely
moves the vasopressor axis and does not reproduce the AI Clinician's tendency toward
more low-dose vasopressor, a difference we cannot attribute to any single change
among the cohort, the evolution of practice, and the variable set.

The agreement figures are most informative conditioned on the clinician's decision
(Section~\ref{sec:agreement}): the policy matches the clinicians far more often when
they withhold treatment than when they administer it. This asymmetry shows that the
policy departs from practice precisely in the harder cases, where treatment is given,
and is a second manifestation of the agreement-intervention circularity we return to
as a limitation (Section~\ref{sec:limitations}). The agreement gap between survivors
and non-survivors should not be over-read as evidence of quality: survivors are the
majority and the more predictable group.

\subsection{The trade-off in state-space size}
\label{sec:ope-tension}
Evaluating with two estimators of different failure modes, rather than the single
weighted importance sampling of the AI Clinician \cite{komorowski2018}, is what
gives the reading its robustness: a good result under one estimator depends
entirely on its own bias and variance, whereas agreement between two whose
weaknesses differ is harder to obtain by artifact. It is also what exposes how
the three quantities move against the number of states $K$, isolated to the
winning variable set and discount (Section~\ref{sec:configuration}): fitted Q
evaluation falls steadily as $K$ grows, from a comfortable margin at
$K=\num{400}$ to a narrower but still positive one at $K=\num{1200}$, never
crossing below the clinicians' return anywhere in the range; weighted importance
sampling crosses from below to above the clinicians' return between
$K=\num{400}$ and $K=\num{600}$ and peaks exactly at $K=\num{1000}$; and the
effective sample size peaks one step earlier, at $K=\num{800}$, already
declining by $K=\num{1000}$. The three therefore do not move together: FQE never
becomes fragile in this range, but WIS and the effective sample size peak at
different points. A plausible interpretation is that a small $K$ yields broad states that merge distinct clinical situations,
giving a misspecified model and an inflated, biased FQE that recedes as $K$
sharpens the states, while WIS keeps gaining resolution until $K=\num{1000}$;
past that point each state retains so little support that too many actions are
masked and the effective sample size has already started to fall, even though it
has not yet crossed the reliability floor at the exact configuration selected.
$K=\num{1000}$ is therefore not the point where every quantity is at its best,
but the compromise where WIS is maximal and the effective sample size, though
past its own peak, still clears the floor that keeps the estimate trustworthy.

That compromise calls for an honest reading of its cost. Both estimators clear the clinicians' return at the final configuration (Section~\ref{sec:ope-results}), on the point estimate and on the lower
confidence bound alike, so the margin itself is not the fragile part of this
result. The fragility sits instead in the effective sample size, which barely
clears its own floor (\num{50.1} against a preset minimum of \num{50}), and in
the discount: every reliable configuration of the sweep sits at the lower
$\gamma=0.95$, none at the $\gamma=0.99$ value the reference work uses. A shorter effective horizon and a
reliability margin close to the floor are therefore the price of this defensible
advantage over the clinicians, not a free result. The sweep underscores how
selective the reliability requirement is: only \num{77} of the \num{1200}
configurations reached an effective sample size of at least 50, all at
$\gamma=0.95$ and from the main variable set, conditions for a reliable estimate
rather than free choices.

Much of that reliability rests on how the clinicians' policy is estimated, the
denominator of the importance ratio that sustains WIS; the control experiment
confirms it, since replacing the random forest with smoothed empirical counts
collapses the effective sample size (Section~\ref{sec:ope-results}). It is worth
being precise about the role of that forest, to avoid a common misreading: it
estimates the clinicians' policy and enters only the evaluation, as the denominator
of the importance ratio, playing no part in building the learned policy, which comes
entirely from policy iteration on the MDP. There is no model steering another, and
the forest would make no sense as a dosing rule, since it describes what clinicians
did without optimizing the outcome. Its errors stay bounded by the smoothing toward
the uniform, the per-step clipping and the monitoring of the effective sample size.
All of this returns to the off-policy evaluation paradox \cite{gottesman2019}: a
more divergent policy holds more potential value but less data to corroborate it,
the balance any offline clinical RL must strike.

This tabular design is a deliberate choice rather than a limitation of ambition.
The aim of this work is not to maximize predictive or clinical performance, which
the retrospective, off-policy setting bounds regardless of model class, but to
keep the evaluation of the learned policy interpretable end to end: every state,
action, and value estimate can be inspected and traced back to the data that
produced it. More expressive alternatives, such as deep reinforcement learning
with function approximation, could plausibly improve the point estimate, but they
would also compound the very interpretability problem this paper addresses, since
function approximation and its diagnostics are harder to audit than a finite
state-action table. We therefore treat those methods as a natural extension once
this more transparent evaluation is established, rather than as a competing
baseline to beat on predictive grounds.

\subsection{Clinical validity and decision support}
\label{sec:clinical-validity}
Distinct from the statistical soundness of the evaluation, this final question is
whether what the policy learned makes clinical sense and holds beyond the particular
sample it was trained on. This is the third of Gottesman's points \cite{gottesman2019}:
prospective behavior and the risk that a policy fails to transfer to another
hospital or time. Several signals support plausibility and consistency, though none
amounts to prospective validation.

The first signal is the cohort itself: its absolute mortality coincides almost
exactly with the AI Clinician's external validation cohort (eRI) and its
demographics and baseline severity are essentially those of the reference study
(Section~\ref{sec:mortality}), so the model is built on a population resembling an
already validated multicenter cohort rather than an atypical profile. This does not
dispel the concern of distribution shift \cite{moreno-torres2012}, since training
still rests on a single center (Section~\ref{sec:limitations}), but it suggests a
clinically representative starting point.The second signal is coherence: two independently constructed models, over different data and variable sets, converge on the same direction of treatment. This supports plausibility, since two models are less likely to share exactly the same artifact than to capture a reproducible clinical signal; as noted, that convergence is firm for fluids and does not extend to the vasopressor axis, and the claim is confined accordingly. A third signal is imposed at
selection rather than found a posteriori: the decision rule required a
non-intervention rate close to the clinicians' (Section~\ref{sec:variable-selection}),
which the final configuration meets, so the policy behaves as a clinical strategy and
not a degenerate optimizer.

A last signal is expressed in observed mortality rather than the estimators' return
scale: grouping validation steps by how far the clinician's dose departs from the
recommendation, in-hospital mortality is lowest at exact agreement and rises as the
two diverge, for both levers (Figure~\ref{fig:agreement-dose}b,c). This is an
observed association, not causal evidence, since the sickest patients tend to receive
the most extreme and hence most divergent treatments; with that reserve, the curve
points in the same direction as the off-policy evaluation without resting on its
assumptions. The counterfactual it raises, how many divergently treated patients
would have survived under the learned policy, is exactly what no observational data
can resolve and what the off-policy estimators approximate.

\begin{figure}[t]
  \centering
  \includegraphics[width=\linewidth]{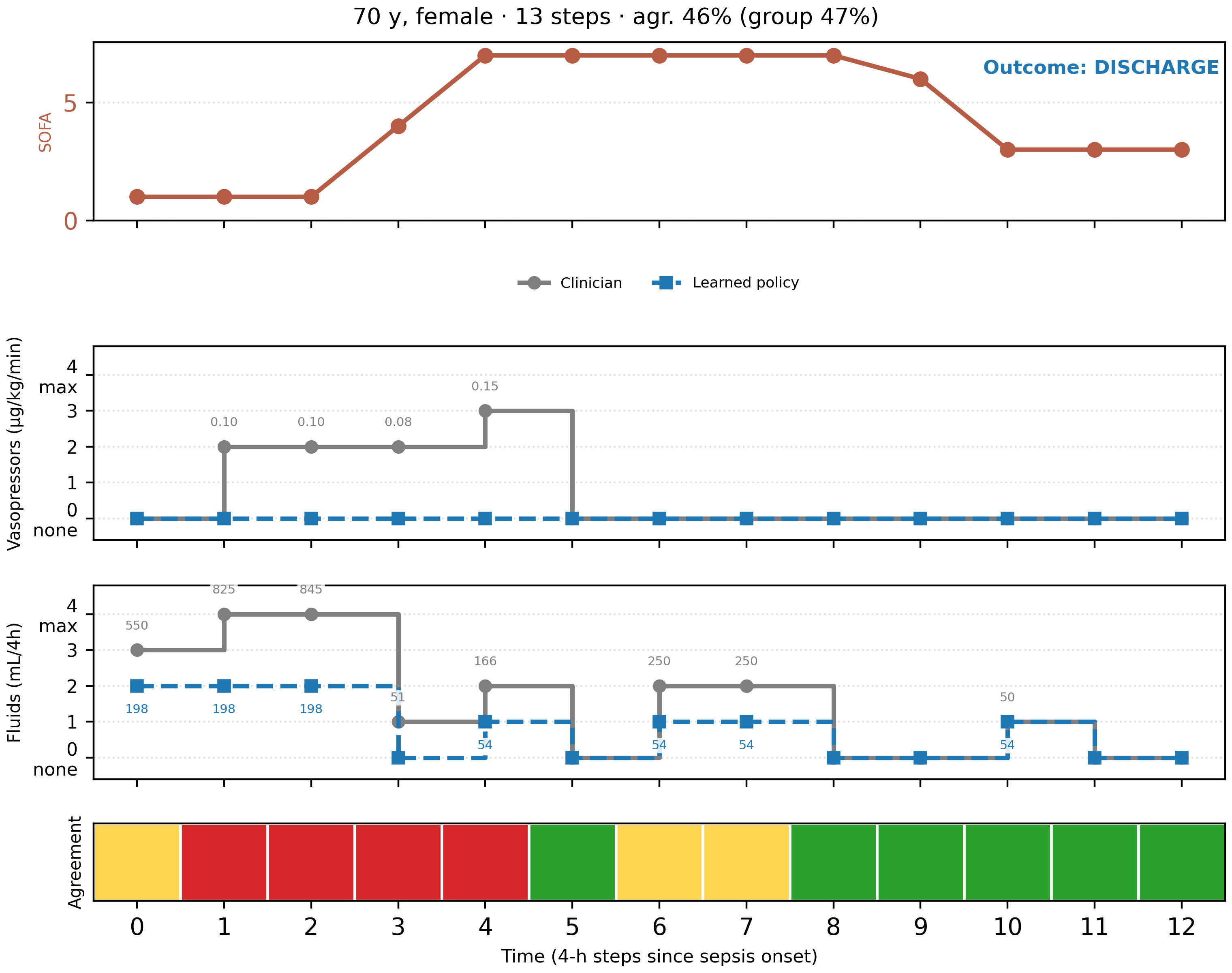}
  \caption{One illustrative discharged patient's trajectory, near the mean agreement
  of her outcome group: SOFA score over the stay; the clinician's and the learned
  policy's vasopressor and fluid doses at each \qty{4}{\hour} step; and a band marking
  step-by-step agreement (exact match, tolerant within one bin, or mismatch).}
  \label{fig:trajectories-main}
\end{figure}

By way of illustration, and without drawing any statistical conclusion,
Figure~\ref{fig:trajectories-main} shows this comparison for one discharged patient
near the mean agreement of her outcome group: the SOFA course, the clinician's fluid
and vasopressor doses against the policy's at each \qty{4}{\hour} step, and a band
marking step by step whether the two agreed. Discrepancy concentrates on the fluid axis while vasopressors barely move, reproducing at the individual scale the between-lever asymmetry seen in aggregate. This example is not inferential evidence, but it gives clinical texture to the aggregate pattern. In this sense, the policy is most naturally interpreted as a basis for discordance-based clinical decision support: highlighting well-supported departures from observed dosing practice rather than prescribing autonomous treatment.

In a future clinical setting, the natural role for such a policy would be decision support: highlighting discordant dosing decisions without replacing clinical judgment. Such a tool would also provide the prospective validation platform that retrospective data cannot supply (Section~\ref{sec:future-work}). Taken together, these signals support the plausibility and consistency of the policy, not a demonstrated clinical benefit: this remains retrospective off-policy evaluation, not prospective validation on patients. The other side of Gottesman's third point, a terminal, sparse reward
that reduces the whole outcome to mortality, is an underlying limitation addressed
next (Section~\ref{sec:limitations}).

%% ---------------------------------------------------------------------
%% 7. LIMITATIONS AND FUTURE WORK
%% ---------------------------------------------------------------------
\section{Limitations and future work}
\label{sec:limitations}
% Source: Cap. 06 Limitaciones + Cap. 07 Trabajo futuro.

\subsection{Limitations}
\label{sec:limitations-list}
The three points of Gottesman et al.\ \cite{gottesman2019} that framed the discussion
each mark a concrete limitation, and we state them in the same order. None is unique
to our model; that generality is why they are posed as challenges for reinforcement
learning in health, and what follows is how each surfaces here, separating what our
implementation introduces from what it inherits from the paradigm and from the
reference work.

\paragraph{State representation and confounding}
The first point concerns whether the state variables faithfully represent the
information on which the clinician acts, and the risk of confounding when they do
not. Here lies the underlying limitation of the work: the block doses, the fluid
volume and the maximum vasopressor rate, enter the state vector of a step and at the
same time define that step's action (Section~\ref{sec:mdp}). They are the
current-step doses at instant $t$, not the previous step's at $t-1$, so the query
state is not prospectively constructible: forming it at the bedside would require
knowing the dose about to be given, and the model as it stands is therefore not prospectively usable without redesigning the state definition. This is not a defect we introduced but a faithful replica of the
AI Clinician design \cite{komorowski2018}, where those block doses are likewise state
variables and the fluid volume is weighted twice as heavily as the rest; its natural
correction, lagging the doses to the previous step, is left as future work
(Section~\ref{sec:future-work}). This circularity is the same one seen at the level of
agreement, where intervention matching is almost perfectly correlated with the
intervention rate itself (Section~\ref{sec:variable-selection}): two readings of one
problem.

To that confounding a different imprecision is added, of our own making rather
than inherited unchanged from the reference work: the SOFA component of the
state is recomputed on the imputed data rather than taken zero-filled from the
derived view (Section~\ref{sec:cohort-char}), which corrects a systematic
downward bias but is not itself exact. The cardiovascular component uses only
the vasopressor rate, since individual-drug doses are unavailable in this form,
and the renal component's 24-hour urine output is approximated by a rolling sum
of six 4-hour blocks, which tends to overstate severity at the very first step.
Because the clinicians' behavior-policy estimator
(Section~\ref{sec:behaviour-policy}) is trained on the same features including
this score, the correction is not confined to the state the policy sees: it
also reaches the denominator of the importance ratio that weighted importance
sampling depends on (Equation~\eqref{eq:ratio}), so the two cannot be varied
independently. Fitted Q evaluation, which does not use the behavior policy at all (Section~\ref{sec:ope}), is unaffected by this particular behavior-policy coupling, although it still depends on the same state representation. A
further inconsistency, smaller but worth stating plainly, is that cohort
selection and sepsis onset (Section~\ref{sec:cohort}) are still detected on the
uncorrected score from the official \texttt{sepsis3} view: only the state
variable and the severity we report are recomputed, not the criterion that
decides who enters the cohort and when their trajectory begins. The
discretization, moreover, admits no clinical novelty: $k$-means
assigns each step to the nearest centroid with no reject option, so a patient
resembling none of the learned groups is absorbed into the least distant state and
inherits its recommendation even when poorly represented. Finally, the whole
formulation rests on the Markov assumption (Section~\ref{sec:mdp}), an approximation
to a strictly partially observable problem.

\paragraph{Fragility of the off-policy evaluation}
The second point is the off-policy evaluation paradox: the more a policy departs from
the clinicians', the more valuable it may be in potential but the less data exist to
corroborate it. Its consequences were examined as the trade-off in state-space size
(Section~\ref{sec:ope-tension}); we record them here only as limits: the effective
sample size barely clears the threshold ($\mathrm{ESS} = \num{50.1}$ against a preset
floor of \num{50}), only \num{77} of the \num{1200} configurations proved reliable at
all, and every one of them at a lower discount than the reference work's, so a
shorter effective horizon and a reliability margin close to the floor are the
price of a defensible margin over the clinicians rather than a free result.
Selecting over \num{1200} configurations also carries a risk of overfitting to
the evaluation criterion itself, analogous to model selection on a validation
set: ranking the 26 configurations that clear the first three gates by their
point margin alone would in fact have favored one whose confidence interval for
WIS barely reaches the clinicians' return, so we added a fourth check requiring
the lower bound of both estimators' intervals, not just their mean, to exceed it
(Section~\ref{sec:ope}), evidence that the point margin alone is not a robust
enough criterion in this part of the sweep. We mitigate the broader risk by
requiring two estimators of differing weaknesses to exceed the clinicians' mark
at once, together with a non-circular plausibility gate
(Section~\ref{sec:variable-selection}), rather than trusting a single number.

This fragility is compounded by a structural limitation of the tabular model:
discretizing the state into $K$ groups and estimating transitions and rewards by
finite counts yields a necessarily misspecified, data-poor model, in which after
support filtering \qty{81.7}{\percent} of state-action pairs lack a reliable
transition, \qty{3.1}{\percent} of states fall back to the clinicians' mode and
\qty{3.2}{\percent} are very small clusters (Section~\ref{sec:configuration}).
Discretization also turns each group into a nominal label and discards the metric
between states, so two clinically neighbouring situations are as disconnected as any
two and each state is estimated in isolation. That scarcity, and the inability to lean
on similar states, are the source of the bias fitted Q evaluation carries and the
reason deep reinforcement learning on the continuous state is raised as future work.

\paragraph{Prospective validity and reward design}
The third point questions prospective validity, along two routes that in our case
become two limitations. The first is the reward design: terminal and sparse, it
reduces the whole outcome to binary in-hospital mortality and forgoes any intermediate
objective or later quality of life, such as post-sepsis syndrome; incorporating richer
intermediate signals is a line we leave open. The second is the risk of distribution
shift: the model is trained on a single center and on a database different from the
reference work (Section~\ref{sec:cohort}), which, despite the cohort's consistency with
the eRI external validation cohort seen above (Section~\ref{sec:clinical-validity}),
leaves its transfer to another hospital or time unguaranteed; variability across data
sources is a documented factor of bias and reduced generalization in clinical machine
learning \cite{saez2021,moreno-torres2012}. A related generalizability question is sex:
although we report the cohort's sex distribution (Section~\ref{sec:cohort-char}), we did
not stratify the policy or its evaluation by sex, so any differential performance across
male and female patients remains unexamined. A minor data limitation adds to this: the
date of death is censored at one year after discharge for anonymization, so
longer-term mortality is unrecorded, though this does not affect the outcomes we use.

Taken together, these limitations do not invalidate the result, but they define its scope: a clinically plausible refinement of observed practice, supported by retrospective off-policy evidence and requiring prospective validation before clinical use.

\subsection{Future work}
\label{sec:future-work}
Each of the three limitations opens a line of work, which we set out in the same order
before closing with a set of lesser methodological refinements.

The most immediate line is to break the state-action circularity by lagging the doses to the previous step $t-1$, so the state describes the patient before the decision and no longer embeds the action about to be taken. This removes the most direct circularity, makes the state constructible at the bedside, and is a prerequisite for prospective decision support. A second line addresses the fragility of the tabular model through deep
reinforcement learning, able to operate on the continuous state without discretizing
\cite{raghu2017} by means of a function approximator such as a deep Q-network
\cite{mnih2015}; the action space could likewise be treated as a continuous dose, which
would call for continuous-action actor-critic methods. Such an approximator might in
principle absorb the variable selection that here required a deliberate sweep, but this
promise should be treated with caution, since our own evidence runs against adding
every variable and letting the model decide: our best results came from restricting the
set to the clinically essential, so automatic selection is worth exploring as a
complement to, not a substitute for, clinical judgment. A third line concerns the
reward: incorporating denser intermediate signals tied to the evolution of markers such
as lactate or SOFA over the stay, or to later quality of life and post-sepsis syndrome,
would guide the policy with a richer signal than the final outcome alone.

Beyond correcting these limitations, the line of widest reach points to deployment. A responsible route toward practice would be a clinical decision-support tool that flags when a clinician's dosing departs from the recommendation without replacing their judgment. That tool would also be the prospective validation platform now lacking, and would allow the model to be updated on the data it generates, with the clinician remaining the actor. Such a loop could ease distribution shift and enrich precisely the regions where data are now scarce: the divergent actions that sustain the evaluation at a low effective sample size. The same loop also carries a risk worth acknowledging, automation bias, if the tool comes to condition the clinician's decision \cite{goddard2012,rajkomar2019}.

Several additional methodological refinements remain. A third, doubly robust off-policy estimator (WDR)
\cite{jiang2016}, combining the virtues of weighted importance sampling and fitted Q
evaluation, would reinforce the triangulation the two current estimators sustain. The
clinicians' policy could be estimated by alternatives to the random forest, such as an
approximate nearest-neighbour method. The state space admits alternatives to $k$-means
with a manual sweep of $K$, such as a Gaussian mixture model \cite{reynolds2009} or
other methods that set the number of groups automatically. And, since sepsis does not
follow a single treatment pattern, a natural line is to stratify the policy by severity
subgroups, for instance by baseline SOFA, rather than learning one policy for the whole
cohort.

%% ---------------------------------------------------------------------
%% 8. CONCLUSIONS
%% ---------------------------------------------------------------------
\section{Conclusions}
\label{sec:conclusions-final}
% Source: Cap. 07.
We revisited the dosing of intravenous fluids and vasopressors in sepsis as an
offline reinforcement learning problem, carrying the AI Clinician framing to the
MIMIC-IV cohort and resolving a discretized Markov decision process by policy
iteration. The main contribution of this work lies less in the learned policy itself than in the discipline of its evaluation: a dual off-policy evaluation that pairs weighted importance sampling with fitted Q evaluation, uses the effective sample size as a reliability diagnostic, and includes clinician agreement as an independent clinical plausibility check, extending the single-estimator evaluation of the reference work.

Two findings stand out. First, the signal-to-noise ratio of the state depends more on the composition of the variable set than on its size: a deliberate selection of clinically essential variables outperformed larger sets, which tended to degenerate toward inaction. Second, both estimators placed the learned policy above the clinicians' return (WIS \num{50.8} and FQE \num{46.8} against \num{38.2}, with $\mathrm{ESS} = \num{50.1}$ above the prespecified reliability floor). This result must be read together with the policy's modest departure from observed practice (total variation \num{0.18}): the learned policy is best interpreted as a clinically plausible refinement of observed care, supporting its interpretation as a clinically plausible refinement of observed care rather than as evidence of a substantially superior or genuinely new treatment strategy.

These results rest on retrospective, single-center, off-policy evidence, and should therefore be read as support for further validation of the policy in a clinical decision-support context, with external validation, temporally cleaner state definitions, and prospective assessment as the next necessary steps.

%% ---------------------------------------------------------------------
%% DECLARATIONS  (see declarations.tex for drafting notes)
%% ---------------------------------------------------------------------
\input{declarations}

%% ---------------------------------------------------------------------
%% BIBLIOGRAPHY
%% ---------------------------------------------------------------------
\bibliography{references}

\end{document}

\maketitle

This Supplementary Material collects the descriptive tables and secondary figures
referenced from the main text, where each is cited in order of appearance.

\FloatBarrier
\clearpage

\section{Supplementary figures}

\begin{figure}[!htbp]
  \centering
  \includegraphics[width=0.72\linewidth]{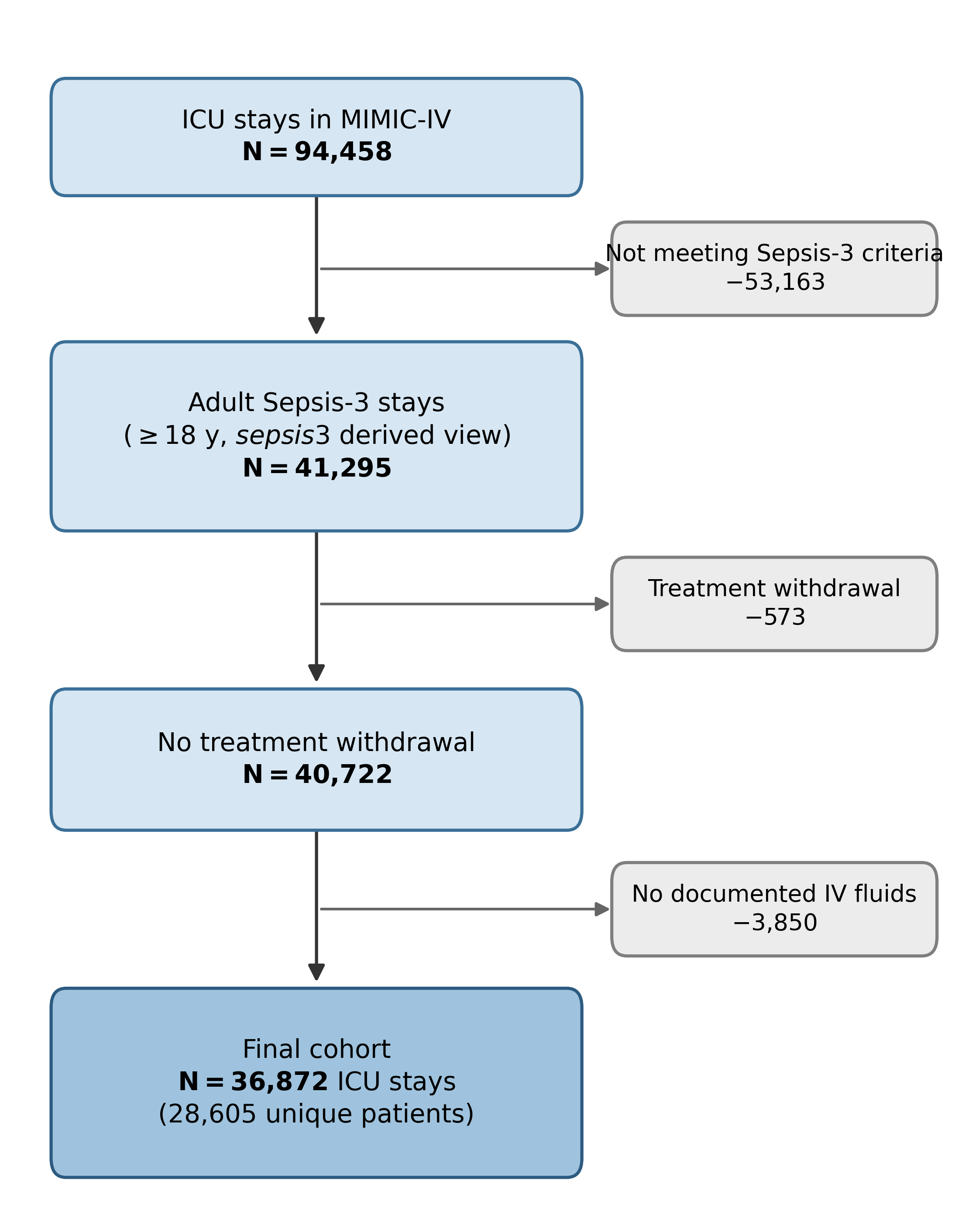}
  \caption{Cohort selection flow. From the \num{94458} ICU stays in MIMIC-IV, the
  adult Sepsis-3 criteria (\texttt{sepsis3} derived view) delimit \num{41295}
  stays, to which two cumulative exclusions are applied, treatment withdrawal and
  the absence of documented intravenous fluids. The final cohort comprises
  \num{36872} ICU stays (\num{28605} unique patients), \qty{89.3}{\percent} of the
  Sepsis-3 stays.}
  \label{fig:consort}
\end{figure}

\begin{figure}[!htbp]
  \centering
  \includegraphics[width=0.82\linewidth]{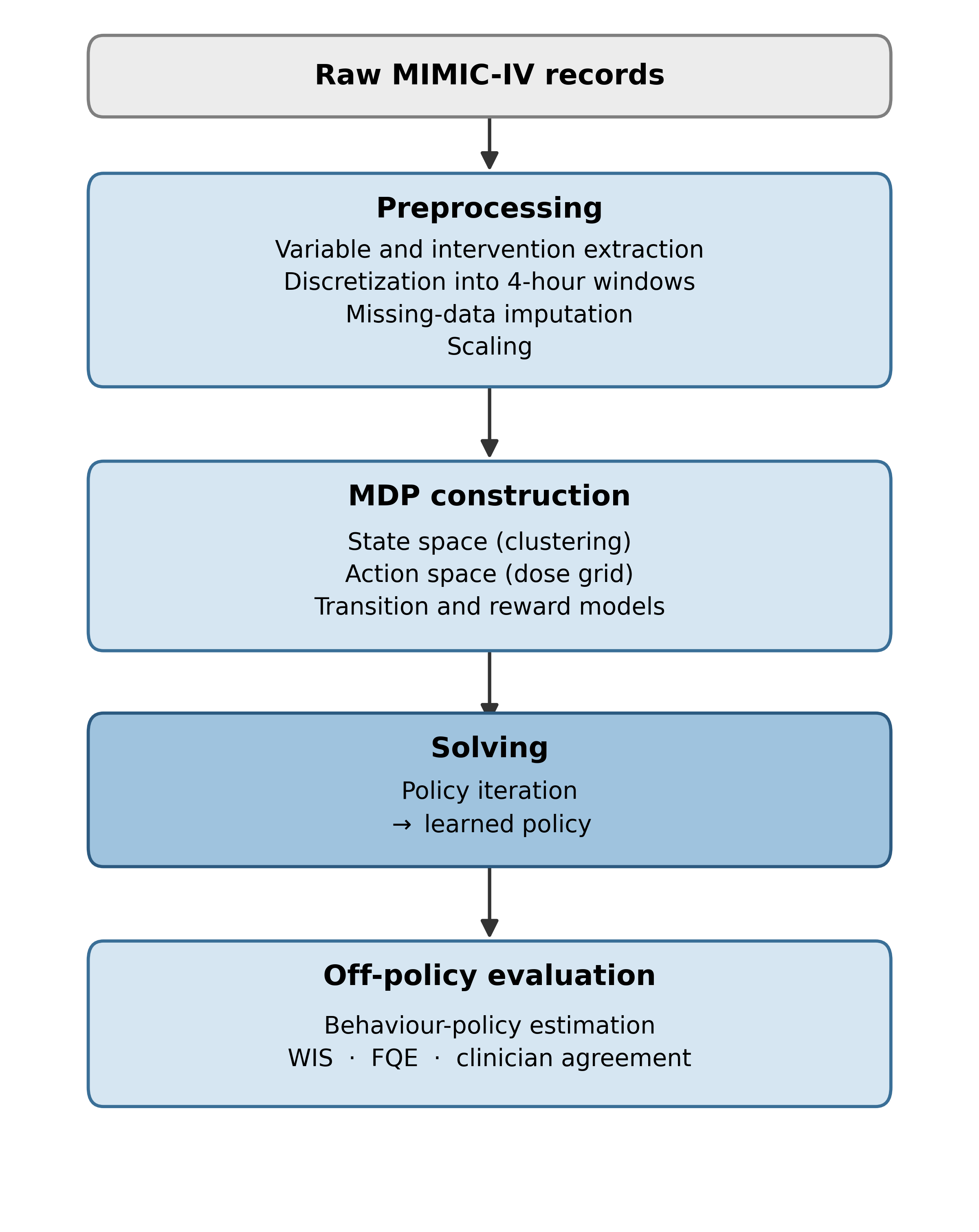}
  \caption{Overview of the preprocessing and MDP-construction pipeline:
  extraction, discretization into \qty{4}{\hour} windows, imputation, scaling,
  state clustering, and estimation of the transition and reward models.}
  \label{fig:pipeline}
\end{figure}

\FloatBarrier
\clearpage

\section{Supplementary tables}

Table~\ref{tab:baseline} reports the clinical state of the cohort in the initial
\qty{4}{\hour} block after sepsis onset, as observed values without imputation.
Coverage is incomplete for several laboratory variables at onset (for example
lactate and total bilirubin are recorded in fewer than half of the patients),
which motivates the imputation strategy of the main text.

\begin{table}[!htbp]
\centering
\small
\caption{Baseline clinical state of the cohort in the initial \qty{4}{\hour}
block $[0, 4)$ after onset. Each variable is summarized as median with
interquartile range [IQR] and as mean $\pm$ standard deviation (SD), computed over
observed values without imputation.}
\label{tab:baseline}
\begin{tabular}{llll}
\toprule
Category & Variable (unit) & Median [IQR] & Mean $\pm$ SD \\
\midrule
\multicolumn{4}{l}{\emph{Vital signs}} \\
 & Heart rate (bpm)                 & 86.2 [75.0, 100.0]  & 88.2 $\pm$ 18.6 \\
 & Systolic blood pressure (mmHg)   & 114.7 [103.8, 128.7]& 117.4 $\pm$ 19.5 \\
 & Mean arterial pressure (mmHg)    & 77.0 [69.2, 86.0]   & 78.5 $\pm$ 13.6 \\
 & Respiratory rate (min$^{-1}$)    & 19.0 [16.0, 22.8]   & 19.8 $\pm$ 5.1 \\
 & Temperature ($^{\circ}$C)        & 36.8 [36.4, 37.2]   & 36.8 $\pm$ 0.9 \\
 & Oxygen saturation (\%)           & 98.0 [95.8, 99.5]   & 97.2 $\pm$ 3.1 \\
 & Glasgow Coma Scale               & 15.0 [14.0, 15.0]   & 14.1 $\pm$ 2.4 \\
\midrule
\multicolumn{4}{l}{\emph{Blood gas}} \\
 & pH                               & 7.4 [7.3, 7.4]      & 7.34 $\pm$ 0.10 \\
 & Lactate (mmol/L)                 & 2.1 [1.4, 3.3]      & 2.8 $\pm$ 2.5 \\
 & PaO\textsubscript{2}/FiO\textsubscript{2} (mmHg) & 231.0 [145.0, 333.0] & 252.8 $\pm$ 151.0 \\
\midrule
\multicolumn{4}{l}{\emph{Chemistry}} \\
 & Creatinine (mg/dL)               & 1.1 [0.8, 1.9]      & 1.7 $\pm$ 1.9 \\
 & Urea nitrogen (mg/dL)            & 24.0 [15.0, 40.0]   & 32.2 $\pm$ 25.7 \\
 & Sodium (mmol/L)                  & 138.0 [135.0, 141.0]& 138.0 $\pm$ 6.2 \\
 & Potassium (mmol/L)               & 4.2 [3.8, 4.8]      & 4.4 $\pm$ 0.9 \\
 & Bicarbonate (mmol/L)             & 23.0 [20.0, 26.0]   & 22.7 $\pm$ 5.3 \\
 & Albumin (g/dL)                   & 3.3 [2.7, 3.7]      & 3.2 $\pm$ 0.7 \\
\midrule
\multicolumn{4}{l}{\emph{Hematology}} \\
 & White blood cells ($\times 10^{3}/\mu$L) & 11.7 [8.0, 16.6] & 13.5 $\pm$ 11.5 \\
 & Hemoglobin (g/dL)                & 10.6 [9.0, 12.3]    & 10.7 $\pm$ 2.3 \\
 & Platelets ($\times 10^{3}/\mu$L) & 190.0 [131.0, 268.0]& 213.3 $\pm$ 125.8 \\
\midrule
\multicolumn{4}{l}{\emph{Hepatic and coagulation}} \\
 & Total bilirubin (mg/dL)          & 0.7 [0.4, 1.6]      & 2.1 $\pm$ 4.7 \\
 & INR                              & 1.3 [1.1, 1.6]      & 1.6 $\pm$ 1.1 \\
\bottomrule
\end{tabular}
\end{table}

\begin{table}[!htbp]
\centering
\caption{State variables of the final configuration (22 variables). The
transformation applied before clustering is coded as: \emph{log},
$\log(1+x)$ followed by standardization; \emph{z}, standardization only;
\emph{bin}, centered binary. The empirical selection of this set is described
in the main-text variable selection.}
\label{tab:state-variables}
\begin{tabular}{lll}
\toprule
Variable & Clinical role & Transf. \\
\midrule
\multicolumn{3}{l}{\emph{Vital signs}} \\
\texttt{heart\_rate}  & Heart rate                    & z \\
\texttt{mbp}          & Mean arterial pressure        & z \\
\texttt{resp\_rate}   & Respiratory rate              & z \\
\texttt{spo2}         & Peripheral oxygen saturation  & z \\
\texttt{temperature}  & Temperature                   & z \\
\midrule
\multicolumn{3}{l}{\emph{Clinical scores}} \\
\texttt{gcs}          & Glasgow Coma Scale            & z \\
\texttt{sofa\_score}  & SOFA (global severity)        & z \\
\texttt{sirs\_score}  & SIRS criteria count           & z \\
\midrule
\multicolumn{3}{l}{\emph{Laboratory}} \\
\texttt{log\_lactate}         & Lactate (perfusion)              & log \\
\texttt{pao2fio2ratio}        & PaO\textsubscript{2}/FiO\textsubscript{2} (respiratory) & z \\
\texttt{log\_creatinine}      & Creatinine (renal)               & log \\
\texttt{log\_bilirubin\_total}& Bilirubin (hepatic)              & log \\
\texttt{platelet}             & Platelets (coagulation)          & z \\
\midrule
\multicolumn{3}{l}{\emph{Demographics and static context}} \\
\texttt{anchor\_age}            & Age                 & z \\
\texttt{is\_male}               & Sex                 & bin \\
\texttt{weight}                 & Body weight         & z \\
\texttt{elixhauser\_vanwalraven}& Comorbidity score   & z \\
\midrule
\multicolumn{3}{l}{\emph{Treatment and balance context}} \\
\texttt{log\_input\_4h}    & IV fluids given in window   & log \\
\texttt{log\_vaso\_rate}   & Vasopressor dose            & log \\
\texttt{log\_urine\_output}& Urine output                & log \\
\texttt{mech\_vent}        & Mechanical ventilation      & bin \\
\texttt{cum\_balance}      & Cumulative fluid balance    & z \\
\bottomrule
\end{tabular}
\end{table}

\begin{table}[!htbp]
\centering
\caption{Action discretization. Each lever is split into five levels by the
quartiles of strictly positive doses on the training split. Vasopressor doses
are norepinephrine equivalents.}
\label{tab:actions}
\resizebox{\columnwidth}{!}{%
\begin{tabular}{llllll}
\toprule
Level & 0 & 1 & 2 & 3 & 4 \\
\midrule
IV fluids per \qty{4}{\hour} (\unit{\milli\litre})
  & $0$ & $(0,100]$ & $(100,259.7]$ & $(259.7,650]$ & $>650$ \\
Vasopressor (\unit{\micro\gram\per\kilogram\per\minute})
  & $0$ & $(0,0.0503]$ & $(0.0503,0.1002]$ & $(0.1002,0.2198]$ & $>0.2198$ \\
\bottomrule
\end{tabular}}
\end{table}

\begin{table}[!htbp]
\centering
\small
\caption{The five candidate variable sets carried into the final sweep. The two
controls (full set and Komorowski replica) reproduce the non-intervention
collapse; the curated sets do not.}
\label{tab:candidate-sets}
\begin{tabular}{llp{4.6cm}l}
\toprule
Set & $n$ & Inclusion rule & Role \\
\midrule
Sepsis-3 core (\texttt{sepsis3\_min}) & 16 & Treatment, Sepsis-3/SOFA axes and core vital signs & Minimal clinical floor \\
Main set (\texttt{core\_hi75}) & 22 & Coverage $\geq\qty{75}{\percent}$, treatment and Sepsis-3 override, no redundancies & Primary candidate \\
Extended labs (\texttt{core\_labs30}) & 29 & Coverage $\geq\qty{30}{\percent}$ (adds imputed labs) and override, no redundancies & Whether sparse labs help \\
Komorowski replica (\texttt{komorowski46}) & 46 & Pool variables present in \cite{komorowski2018} & Replicative control \\
Full set (\texttt{all50}) & 50 & All pool variables & Negative control \\
\bottomrule
\end{tabular}
\end{table}

\FloatBarrier
\clearpage

\bibliography{references}

%% file: declarations.tex
%% =====================================================================
%%  DECLARATIONS  (AIME / Elsevier required blocks)
%%  SKELETON — content to be confirmed with author/tutor before submission.
%%  Placed before the reference list, per the guide.
%% =====================================================================

\section*{Ethics statement}
This study uses MIMIC-IV, a publicly available, deidentified critical-care database.
The collection of the database was approved by the institutional review boards of the
Beth Israel Deaconess Medical Center and of the Massachusetts Institute of Technology,
which granted a waiver of informed consent; secondary analyses of the deidentified data
are exempt from further review. Access was obtained through the credentialed PhysioNet
process, comprising human-subjects research training and a signed data use agreement.

\section*{Data availability}
The data that support the findings of this study are available from MIMIC-IV, a public
database hosted on PhysioNet, but restrictions apply: they cannot be redistributed by
the authors and require credentialed access and a signed data use agreement. The
database is available at PhysioNet \cite{johnson2023,pollard2026}. The analysis code
is publicly available at
\url{https://github.com/marc-perez-dev/sepsis-rl-mimiciv} (repository
\texttt{mimiciv-sepsis-ai-clinician}).

\section*{Declaration of competing interests}
The authors declare no competing interests.

\section*{Funding}
This research has been partially funded by Agencia Estatal de Investigaci\'on---Proyectos
de Generaci\'on de Conocimiento 2022, project KINEMAI (PID2022-138636OA-I00).

\section*{CRediT authorship contribution statement}
\textbf{Marc P\'erez Roig:} Conceptualization, Methodology, Software, Formal analysis,
Investigation, Data curation, Visualization, Writing -- original draft.
\textbf{David Fern\'andez-Narro:} Supervision, Writing -- review \& editing.
\textbf{Carlos S\'aez:} Conceptualization, Methodology, Supervision,
Writing -- review \& editing.

\section*{Declaration of generative AI and AI-assisted technologies}
During the preparation of this work the authors used a large language model
to assist them reviewing the drafted text and editing the English prose of the manuscript. They take full
responsibility for the content of the published article.

\section*{Acknowledgements}
% TODO (tutor input): language help, compute resources, and people to acknowledge.